\documentclass[10pt,twocolumn,letterpaper]{article}
\usepackage{nopageno}

\usepackage{template/CVPR/cvpr}
\usepackage{times}
\usepackage{epsfig}
\usepackage{graphicx}
\usepackage{amsmath}
\usepackage{amssymb}

\usepackage{multirow}
\usepackage[percent]{overpic}
\usepackage[export]{adjustbox}
\usepackage{comment}

\usepackage{booktabs} 
\usepackage{subcaption}
\usepackage{esint}

\usepackage[pagebackref=true,breaklinks=true,letterpaper=true,colorlinks,bookmarks=false]{hyperref}

\ifcvprfinal\fi

\def\cvprPaperID{8458} 
\cvprfinalcopy 
\begin{document}
\title{NeX: Real-time View Synthesis with Neural Basis Expansion}
%

\author{Suttisak Wizadwongsa\footnotemark[1]\and Pakkapon Phongthawee\footnotemark[1]\and Jiraphon Yenphraphai\footnotemark[1]\and Supasorn Suwajanakorn \\

VISTEC, Thailand\\
{\tt\small \{suttisak.w\_s19, pakkapon.p\_s19,  jiraphony\_pro, supasorn.s\}@vistec.ac.th}
 
}




\twocolumn[{
\renewcommand\twocolumn[1][]{#1}
\maketitle
\vspace*{-10mm}

\begin{center}
    \label{mainmap}
    \centering
    \includegraphics[width=1.025\textwidth]{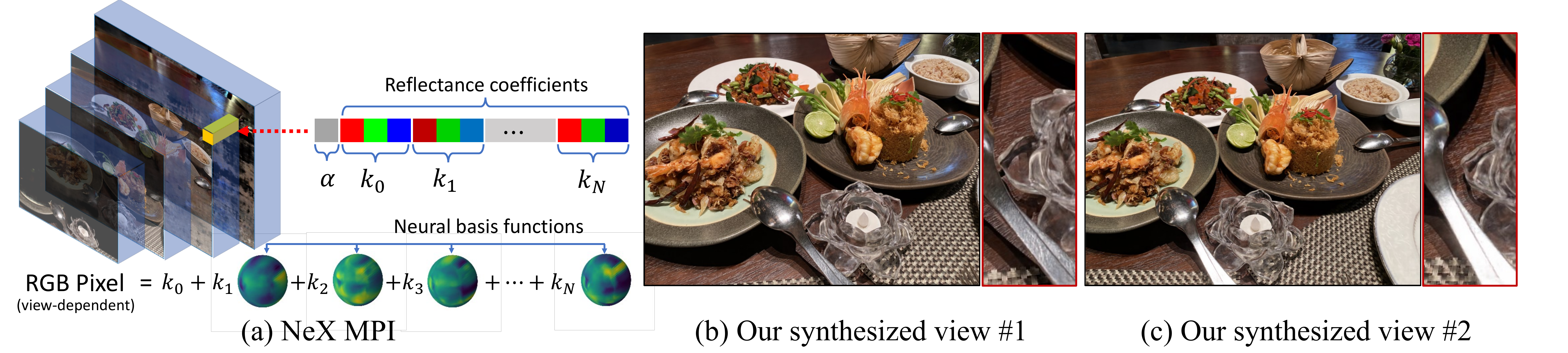}
    \vspace*{-5mm}
    \captionof{figure}{
    \textbf{(a)} Each pixel in NeX multiplane image consists of an alpha transparency value, base color $k_0$, and view-dependent reflectance coefficients $k_1...k_n$. A linear combination of these coefficients and basis functions learned from a neural network produces the final color value.
    \textbf{(b, c)} 
    show our synthesized images that can be rendered in real time with view-dependent effects such as the reflection on the silver spoon. 
    }
\end{center}
}]

\begin{abstract}


We present NeX, a new approach to novel view synthesis based on enhancements of multiplane image (MPI) that can reproduce next-level view-dependent effects---in real time. Unlike traditional MPI that uses a set of simple RGB$\alpha$ planes, our technique models view-dependent effects by instead parameterizing each pixel as a linear combination of basis functions learned from a neural network. Moreover, we propose a hybrid implicit-explicit modeling strategy that improves upon fine detail and produces state-of-the-art results. Our method is evaluated on benchmark forward-facing datasets as well as our newly-introduced dataset designed to test the limit of view-dependent modeling with significantly more challenging effects such as the rainbow reflections on a CD. Our method achieves the best overall scores across all major metrics on these datasets with more than 1000$\times$ faster rendering time than the state of the art. For real-time demos, visit \emph{\smaller \url{  https://nex-mpi.github.io/}}


\end{abstract}
\renewcommand{\thefootnote}{\fnsymbol{footnote}}
\footnotetext[1]{Authors contributed equally to this work.}
\setlength{\belowcaptionskip}{-10pt}

\section{Introduction}
Novel view synthesis is an exciting and long-standing problem that draws much attention from both the computer vision and graphics communities. The problem comprises two intriguing challenges of how to construct a visual scene representation from only a sparse set of images and how to render such a representation from unseen perspectives. A wide range of applications are possible from this area of research ranging from virtually visiting tourist attractions to viewing any online product all around in 3D; however, such experiences would only become most compelling and practical when the representation allows photo-realistic and real-time synthesis.

One candidate that can serve this purpose is multiplane image (MPI) \cite{zhou2018stereo} which approximates the scene's light field with a set of parallel semi-transparent planes placed along a reference viewing frustum. This representation is shown to be more effective than traditional 3D mesh reconstruction in reproducing complex scenes with challenging occlusions, thin structures, or planar reflections. However, the standard RGB$\alpha$ representation of MPI is limited to diffuse surfaces whose appearance stays constant regardless of the viewing angle. This greatly limits the types of objects and scenes that MPI can capture. Recent research on implicit scene representation has made significant progress in the past months \cite{mildenhall2019local,sitzmann2020implicit,liu2020neural,zhang2020nerf++,trevithick2020grf} and can be applied to view synthesis problem. Unfortunately, its expensive network inference still prohibits real-time rendering, and reproducing complex surface reflectance with high fidelity still remains a challenge. Our method breaks these limits on both fronts.

We introduce NeX, a new scene representation based on MPI that models view-dependent effects by performing basis expansion on the pixel representation in our MPI. In particular, rather than storing static color values as in traditional MPI, we represent each color as a function of the viewing angle and approximate this function using a linear combination of spherical basis functions learned from a neural network. Furthermore, we propose a hybrid parameter modeling strategy that models high-frequency detail in an explicit structure within an implicit MPI modeling framework. This strategy helps improve fine detail that is difficult to model by a neural network and produces sharper results in fewer training iterations. 

We evaluate our algorithm on benchmark forward-facing datasets and compare against state-of-the-art approaches including NeRF \cite {mildenhall2020nerf} and DeepView \cite {flynn2019deepview}. These datasets, however, contain mostly diffuse scenes and fairly simple view-dependent effects and cannot be used to judge the new limit of our algorithm. Thus, we collect a new dataset, \emph{Shiny}, with significantly more challenging view-dependent effects such as the rainbow reflections on a CD, refraction through non-planar glassware or a magnifying glass. Our method achieves the best overall scores across all major metrics on these datasets. We provide quantitative and qualitative results and ablation studies to justify our main technical contributions. Compared to the recent state of the art, NeRF \cite {mildenhall2020nerf}, our method captures more accurate view-dependent effects and produces sharper results---all in real time.

  \section{Related Work}
%

\textbf{Learning MPIs.}
Multiplane image by Zhou et al. \cite{zhou2018stereo} is a scene representation that consists of parallel semi-transparent planes placed along a reference viewing frustum. Note that a similar representation has been proposed earlier by the name of ``stack of acetates'' by Szeliski \& Golland \cite{szeliski1998stereo}.  
Originally, MPI~\cite{zhou2018stereo} is used to solve a small-baseline stereo problem and is inferred with a convolutional neural network (CNN) from an input stereo pair.
Subsequent work extends MPI to support multiple input photos \cite{flynn2019deepview,mildenhall2019local,li2020crowdsampling} or even infers an MPI from a single image \cite{tucker2020single}. 
In \cite{mildenhall2019local}, a CNN is used to predict multiple nearby MPIs which are then blended together to produce the final output. Srinivasan et al. \cite{srinivasan2019pushing} predicts an MPI using a two-step process that combines 3D CNNs for MPI prediction and a 2D flow field for warping RGB values from an intermediate rendering. In contrast, \mbox{DeepView}~\cite{flynn2019deepview} uses a CNN to learn gradient updates to the MPI instead of predicting it directly. This learned gradient descent helps avoid over-fitting and requires only a few iterations to generate an MPI. Recently, DeepMPI~\cite{li2020crowdsampling} has been introduced to model time-varying scene appearance and can manipulate colors on their MPI using a CNN. However, these approaches do not model view-dependent effects or only handle them indirectly by blending multiple view-\emph{independent} MPIs. This greatly limits the types of applicable objects and scenes.



\textbf{View synthesis and interpolation.}
One way to categorize view synthesis algorithms is by how dense the input scene is sampled. When the capture is dense as in lumigraph \cite{gortler1996lumigraph,buehler2001unstructured} and light field rendering  \cite{levoy1996light, zhang2003spectral, levin2010linear, Shi2014Light}, the challenge becomes how to store, interpolate, and compress the light field samples.  
When there are only 1-2 input images, the challenge becomes how to infer the ill-constrained 3D geometry and disoccluded regions \cite{srinivasan2017learning, niklaus20193d, tucker2020single,wiles2019synsin, choi2019extreme}. Our work focuses on the case with a moderate number of captures facing forward. 
Besides MPI-based approaches, other solutions include methods based on layered depth images \cite{shade1998layered,tulsiani2018layer,dhamo2019peeking},
Soft3D by Penner et al. \cite{penner2017soft}, which combines depth estimation with soft blending of an estimated geometry, and other 3D reconstruction based techniques \cite{zitnick2004high,hedman2017casual}. 

Neural approaches to view synthesis include DeepStereo \cite{flynn2016deepstereo}, which uses a CNN to predict pixels directly for individual viewing angles and Neural Textures \cite{thies2019deferred}, which combines a reconstructed 3D mesh with neural textures that can be rendered with a neural network. Similar ideas of using neural latent code stored in some geometric structure such as a voxel grid or volumes have been proposed \cite{sitzmann2019deepvoxels, eslami2018neural, lombardi2019neural}. Neural BTF \cite{Rainer2019Neural} represents the bidirectional texture function with an encoder-decoder network that takes in the light and viewing angles and outputs each pixel's color. \cite{kuznetsov2019learning} uses a generative adversarial network to model spatially varying BRDFs of specular microstructures.

One recent notable work is Neural Radiance Fields (NeRF) by Mildenhall et al. \cite{mildenhall2020nerf}, which represents a 5D radiance field with a multilayer perceptron (MLP) that directly regresses the volume density and RGB colors. This method can handle view-dependent effects as the viewing angle is part of the 5D radiance function.
Subsequent work improves upon NeRF by using explicit sparse voxel representation to improve fine detail (NSVF) \cite{liu2020neural}, parameterizing the space to better support unbounded scenes (NeRF++) \cite{zhang2020nerf++}, incorporating learned 2D features that help enforce multiview consistency (GRF) \cite{trevithick2020grf}, or extending NeRF to handle photometric variations and transient objects in internet photo collections (NeRF-W) \cite{martin2020nerf}. Another related line of work involves implicitly modeling  surface reflectance properties in addition to the scene geometry \cite{bi2020neural} or the light transport function \cite{zhang2020neural}. Our work is inspired by these implicit neural representations as well as deep image prior \cite{ulyanov2018deep}, but our goal is directed toward a representation amendable to discretization and real-time rendering. 

\textbf{Light field factorization.}
Our reparameterization of pixel into a combination of basis functions is closely related to light field factorization approaches in many areas, such as surface light field \cite{daniel2000sfl}, precomputed radiance transfer \cite{sloan2002precomputed, sloan2003clustered}, BRDF estimations \cite{kautz1999hardware}, light field and tensor display \cite{wetzstein2012tensor}. In particular, our MPI pixel can be considered generally as a discretized sample of a radiance function $f(\hat{x},\hat{v})$ of position $\hat{x}$ in space and viewing direction $\hat{v}$. This function has been approximated with a sum of products of functions $f(\hat{x},\hat{v})\approx\sum k_n(\hat{x})h_n(\hat{v})$ with techniques such as singular value decomposition (SVD) \cite{kautz1999hardware} or normalized decomposition (ND) \cite{kautz1999interactive}. In precomputed radiance transfer, part of the rendering equation can be similarly formulated as a product of spherical harmonics coefficients of the lighting and transfer function \cite{sloan2002precomputed,ramamoorthi2001efficient}. Clustered PCA \cite{sloan2003clustered} further divides the transfer function matrix into clusters, which are then low-rank-approximated with PCA to reduce rendering cost. A key difference between our method and others is that we model $h_n$ and $k_n$ with neural networks and solve the factorization through network training.




  \section{Approach}
Given a set of multiview images of a scene, our goal is to construct a 3D representation that can render novel views with view-dependent effects in real time. To solve this, we propose a novel representation based on multiplane image \cite{zhou2018stereo} but with significant improvements which include a novel view-dependent pixel representation that can handle non-Lambertian surfaces and a hybrid implicit-explicit parameter modeling to improve fine detail. Our approach focuses on forward-facing captures with around 12 images or more, such as those taken casually with a smartphone. In the following sections, we first briefly review the original MPI representation, then explain our novel representation and a learning method for inferring it.

\begin{figure}
 \centering
  \includegraphics[width=8.2cm]{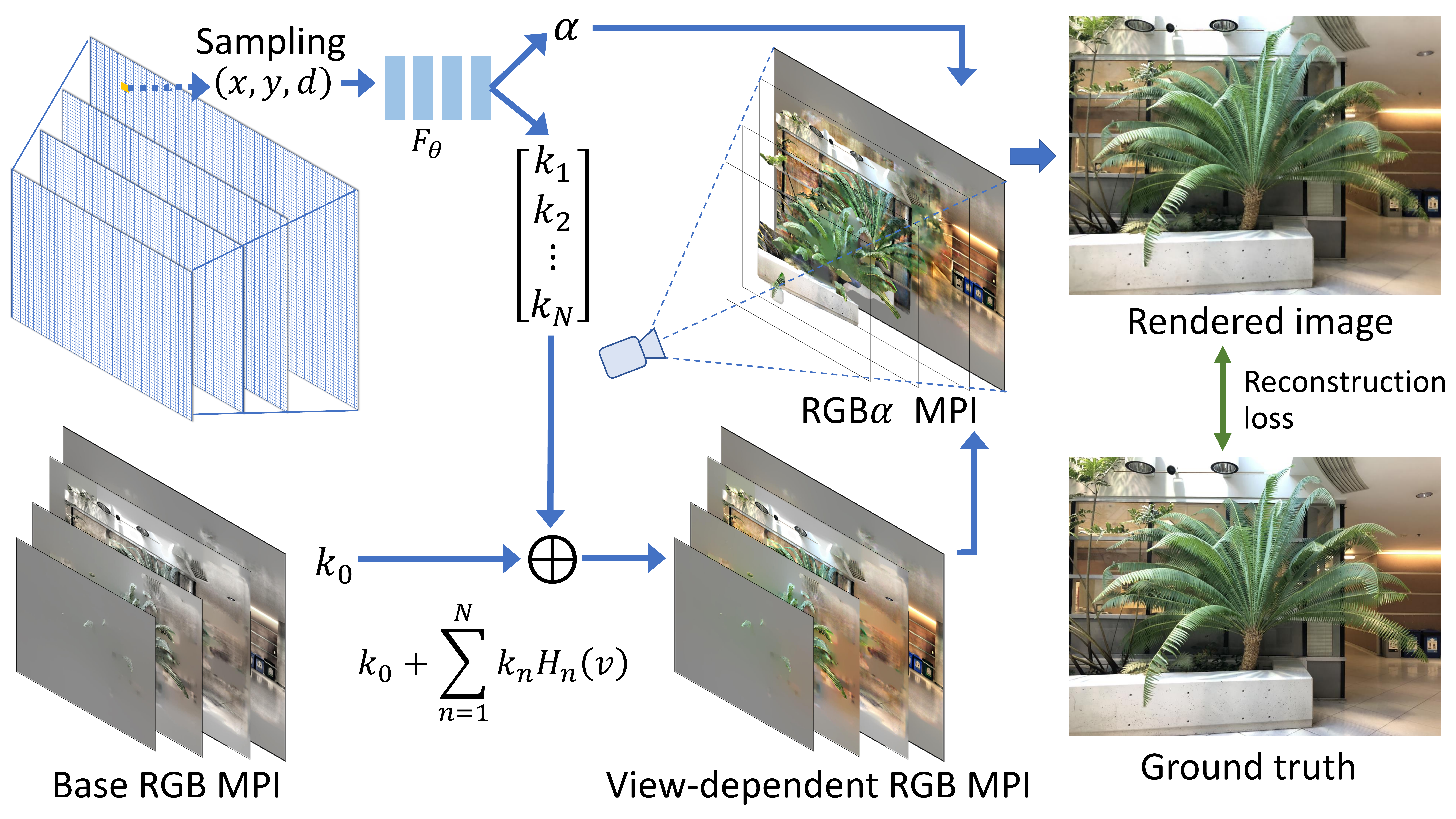}
  \caption{NeX overview: we construct each pixel in our MPI by sampling a pixel coordinate $(x,y)$ at plane depth $d$ and feed it to a multilayer perceptron (MLP) to output alpha transparency and view-dependent basis coefficients $(k_1, k_2,...,k_n)$. These coefficients, together with an explicit $k_0$, are multiplied with basis functions predicted from another MLP, to produce the RGB value. The output image is the product of the composite operation over all planes (Eq. \ref{eq:1}). We train the two MLPs and optimize for the explicit $k_0$ by comparing the rendered image to the ground truth.
  }
  \label{fig:overview}
\end{figure}

\subsection{Original MPI Representation}
Multiplane image \cite{zhou2018stereo} is a 3D scene representation that consists of a collection of $D$ planar images, each with dimension $H\times W\times 4$ where the last dimension contains RGB values and alpha transparency values. These planes are scaled and placed equidistantly either in the depth space (for bounded close-up objects) or inverse depth space (for scenes that extend out to infinity) along a reference viewing frustum (see Figure \ref{fig:overview}).

Rendering an RGB$\alpha$ MPI in any target view can be done by first warping all its planes to the target view via a homography that relates the reference and target view and apply the composite operator \cite{Porter1984Comp}. In particular, let $c_i\in \mathbb{R}^{H\times W \times 3}$ and $\alpha_i\in \mathbb{R}^{H\times W \times 1}$ be the RGB and alpha ``images'' of the $i^\text{th}$ plane, ordered from back to front. And denote $A=\{\alpha_1,\alpha_2,...,\alpha_D\}$, $C=\{c_1,c_2,...,c_D\}$ as the sets of these images. This MPI can then be rendered in a new view, $\hat{I}$, using the composite operator $O$:
\begin{equation}\label{eq:1}
    \hat{I} = O(W(A),W(C))
\end{equation}
where $W$ is a homography warping function that warps each image to the target view, and $O$ has the form:
\begin{equation}\label{eq:2}
    O(A,C) = \sum^{D}_{d=1}c_d T_d(A), \;  T_d(A) = \alpha_d \prod^{D}_{i=d+1} (1-\alpha_i)
\end{equation}
This rendering equation is completely differentiable, thus allowing MPI to be inferred through image reconstruction loss \cite{zhou2018stereo,flynn2019deepview}.

\subsection{View-Dependent Pixel Representation}
One main limitation of MPI is that it can only model diffuse or Lambertian surfaces, whose colors appear constant regardless of the viewing angle.\footnote{A single MPI can simulate planar reflections to some extent by placing the reflected content on one of its planes \cite{flynn2019deepview}.} In real-world scenes, many objects are non-Lambertian, such as a ceramic plate and a glass table. These objects exhibit view-dependent effects such as reflection and refraction. Reconstructing these objects with an MPI can make the objects appear unrealistically dull without reflections or even break down completely (Figure \ref{No_view}) due to the violation of the brightness constancy assumption used for matching invariant and 3D reconstruction. \cite{mildenhall2019local} attempts to solve this by combining multiple view-\emph{independent} MPIs, but their results contain warping artifacts when blending between MPIs.

To allow for view-dependent modeling in our MPI, we modify the pixel color representation, originally stored as RGB values, by parameterizing each color value as a function of the viewing direction $\mathbf{v}=(v_x, v_y, v_z)$. This results in a 3-dimensional mapping function $\mathcal{C}(\mathbf{v}): \mathbb{R}^3 \rightarrow \mathbb{R}^3$ for every pixel. However, storing this mapping explicitly is prohibitive and not generalizable to unobserved angles. Regressing the color directly from $\mathbf{v}$ (and the pixel location) with a neural network, as is done in e.g. \cite{mildenhall2020nerf}, is possible though inefficient for real-time rendering. Our key idea is to approximate this function with a linear combination of \emph{learnable} basis functions $\{H_n(\mathbf{v}): \mathbb{R}^3 \rightarrow \mathbb{R}\}$ over the spherical domain described by vector $\mathbf{v}$:

\begin{equation} \label{eq:basis}
     \mathcal{C}^\mathbf{p}(\mathbf{v}) =  k^\mathbf{p}_0+\sum^{N}_{n=1} k^\mathbf{p}_n H_n(\mathbf{v}) 
\end{equation}
where $k^\mathbf{p}_n \in \mathbb{R}^3$ for pixel $\mathbf{p}$ are RGB coefficients, or reflectance parameters, of $N$ global basis functions. In general, there are several ways to define a suitable set of basis functions. Spherical harmonics basis is one common choice used heavily in computer graphics to model complex reflectance properties. Fourier's basis or Taylor's basis can also be used. However, one shortcoming of these ``fixed'' basis functions is that in order to capture high-frequency changes within a narrow viewing angle, such as sharp specular highlights, the number of required basis functions can be very high. This in turns requires more reflectance parameters which make both learning these parameters and rendering more difficult. With learnable basis functions, our modified MPI outperforms other versions with alternative basis functions that use the same number of coefficients shown in our experiment in Section \ref{typeofbasis}.

To summarize, our modified MPI contains the following parameters per pixel: $\alpha, k_0, k_1, \ldots, k_N$; and global basis functions $H_1, H_2,\dots, H_N$ shared across all pixels.


\subsection{Modeling MPI with Neural Networks} 
Given our modified MPI and the differentiable rendering equation (Eq. \ref{eq:1}), one can directly optimize for its parameters that best reproduce the training views. 
However, as demonstrated in earlier work \cite{flynn2019deepview}, doing so would lead to a noisy MPI that overfits the training views and fails to generalize. 
We can overcome this problem by leveraging the idea of deep prior \cite{ulyanov2018deep} and regressing these parameters with multilayer perceptrons (MLPs) from spatial conditioning, i.e., pixel coordinates. In other words, instead of allowing the estimated parameters to take arbitrary values which are prone to overfitting, we regularize these parameters by only allowing them to take on certain values that are in the span of a deep neural network's output. In our case, we use two separate MLPs; one for predicting per-pixel parameters given the pixel location, and the other for predicting all global basis functions given the viewing angle. The motivation for using the second network is to ensure that the prediction of the basis functions, which are global, is not a function of the pixel location.

Our first MLP is modeled as $F_\theta$ with parameter $\theta$:
\begin{equation}
    F_\theta: (\mathbf{x}) \rightarrow (\alpha, k_1, k_2,...,k_N)
\end{equation}
where $\mathbf{x}=(x,y,d)$ contains the location information of pixel $(x, y)$ at plane $d$. Note that $k_0$ is not predicted by $F_\theta$ but will be stored explicitly in our implicit-explicit modeling strategy, explained in the upcoming section. The second network is modeled as $G_\phi$ with parameter $\phi$:
\begin{equation}
    G_\phi: (\mathbf{v}) \rightarrow (H_1, H_2, ..., H_N)
\end{equation}
where $\mathbf{v}$ is the normalized viewing direction. 

It is interesting to note that in a study from \cite{ulyanov2018deep}, when a CNN is used as a deep prior for synthesizing images, the span of the CNN can capture the natural image manifold surprisingly well. In our case, we found that deep priors based on multilayer perceptrons can regularize our MPI and produce superior results compared to direct optimization without deep priors or with standard regularizers, such as total variation. In relation to NeRF~\cite{mildenhall2020nerf}, our MPI can be thought of as a discretized sampling of an implicit radiance field function that replaces the general view-dependent modeling, predicted with an MLP in NeRF, with more efficient basis functions.


\subsection{Implicit-Explicit Modeling Strategy} 
One observation when using an MLP to model $k_n$, or ``coefficient images'' when $k_n^\mathbf{p}$ is evaluated on all pixels $\mathbf{p}$, is the absence of fine detail (similar reports in \cite{sitzmann2020implicit,mildenhall2020nerf,tancik2020fourier}). In our problem, fine detail or high-frequency content tends to come from the surface texture itself and not necessarily from a complex scene geometry. Thus, we use positional encoding proposed in \cite{mildenhall2020nerf} to regress these images, which helps to an extent but still produces blurry results. Interestingly, we found that simply storing the first coefficient $k_0$, or ``base color,'' explicitly helps ease the network's burden of compressing and reproducing detail and leads to sharper results, also in fewer iterations. With this implicit-explicit modeling strategy, we predict every parameter with MLPs except $k_0$ which will be optimized explicitly as a learnable parameter with a total variation regularizer.

\textbf{Coefficient Sharing:} In practice, computing and storing all $N+1$ coefficients for all pixels for all $D$ planes can be expensive for both training and rendering. In our experiment, we use a coefficient sharing scheme where every $M$ planes will share the same coefficients, but not the alphas. That is, there is a single set of $\{K_0, ..., K_N\}$ for planes $1$ to $M$, and another set for planes $M+1$ to $2M$, and so forth. With proper $N$ and $M>1$, we do not observe any significant degradation in the visual quality, but a significant gain in speed and model compactness.

Finally, to optimize our model, we evaluate the two MLPs to obtain the implicit parameters, render an output image $\hat{I_i}$, and compare it to the ground-truth image $I_i$ from the same view. 
We use the following reconstruction loss:
\begin{equation}
L_{\text{rec}}(\hat{I}_i, I_i) = \| \hat{I}_i- I_i \|^2 + \omega \| \nabla \hat{I}_i-\nabla I_i \|_1,
\end{equation}
where $\nabla$ denotes the gradient operator and $\omega$ is a balancing weight \cite{hqdeblurring_siggraph2008}.
Our approach is summarized in Algorithm 1.

\begin{algorithm}
\small
\label{algo}
\SetAlgoLined
 initialize: $\theta, \phi, K_0$\;
 pre-compute $\mathcal{X}$ pixel coordinate for each pixel\; 
 \For{Iteration=0 to maxIter}{
    sampling image $I_i$\;
    compute $ (A,\vec{K}) = F_\theta(\mathcal{X})  $ where $\vec{K} = [K_1, K_2,...,K_N] $\;
    compute viewing direction $\mathcal{V}_i$ by
     $\mathcal{V}_i = \mathcal{X}- \text{center of projection of} \:I_i; \: \mathcal{V}_i = \mathcal{V}_i / \| \mathcal{V}_i\| $
    \;
    compute view-dependent color $ C = K_0 + \vec{K} \cdot \vec{H}_\phi(\mathcal{V}_i)$\;
    compute rendered image \quad\quad $\hat{I}_i = O(W_i(A),W_i(C)) $ \;
    compute loss function by   $ L = L_{\text{rec}}(\hat{I}_i, I_i) +\gamma \text{TV}(K_0) $ \;
    update $\theta, \phi, K_0$ with ADAM$(\nabla_{\theta,\phi,K_0}L)$ \;
    }
   \KwResult{$A,K_0,K_1,...,K_N$} 
 \caption{MPI training with NeX}
\end{algorithm}

\subsection{Real-time Rendering} 
Every model parameter in our MPI can be converted to an image. This is done by evaluating $F_\theta$ on all pixel coordinates and $G_\phi$ on some pre-defined viewing span. Given these pre-computed images, we can implement Equation \ref{eq:1} in a fragment shader in OpenGL/WebGL and achieve real-time view-dependent rendering of our captured scenes.

\section{Experiments}
We perform quantitative and qualitative evaluations against state-of-the-art methods for novel view synthesis which include MPI-based methods and others. We also provide an extensive study on the choice of the basis functions and evaluate different variations of implicit-explicit modeling of the MPI parameters, ranging from fully implicit to fully explicit. 

\subsection{Implementation Details}
Our model is optimized independently for each scene. The input photos are first calibrated and undistorted with a structure-from-motion algorithm from COLMAP \cite{schoenberger2016sfm}.
In most of our experiments unless stated otherwise, we use an MPI with 192 layers with $M=12$ consecutive planes sharing one set of texture coefficients.

\textbf{MLP architectures:} 
For $F_\theta$ that predicts per-pixel parameters given the pixel location $(x, y, d)$, we follow NeRF's \cite{mildenhall2020nerf} positional encoding and project input $x, y$ to 20 dimensions each and plane depth $d$ to 16 dimensions with the following projection $p(u) = \left[\sin(2^0\frac{\pi}{2}u),\cos(2^0\frac{\pi}{2}u),...,\sin(2^k\frac{\pi}{2}u),\cos(2^k\frac{\pi}{2}u)\right]$ where input $u$ is first normalized to $[-1, 1]$. The total input dimension is 56. This network uses 6 fully-connected LeakyReLU layers, each with 384 hidden nodes. The output $\alpha$ uses a sigmoid activation, and the others use tanh activations. 
For $G_\phi$ that predicts the basis functions, we use positional encoding of the input viewing direction with 12 dimensions including 6 dimensions for $v_x$ and $v_y$. This network uses 3 fully-connected LeakyReLU layers with 64 hidden nodes to output 8 dimensions of $\vec{H}_\phi(\mathbf{v})$. 

\textbf{Training details:} 
To compute the loss, we randomly sample and render 8,000 pixels in the training view and compare them to the corresponding pixels in the ground-truth image.
We set $\omega=0.05$, $\gamma=0.03$ and train our networks for 4,000 epochs using Adam optimizer \cite{Kingma2015AdamAM} with a learning rate of 0.01 for base color and 0.001 for both networks and a decay factor of 0.1 every 1,333 epochs.

\textbf{Runtime:}
For a scene with 17 input photos of resolution 1008 $\times$ 756, the training took around 18 hours using a single NVIDIA V100 with a batch size of 1. Our WebGL viewer can render this scene at 300 frames per second using an NVIDIA RTX 2080Ti. For comparison, NeRF took about 55 seconds to generate one frame on the same machine. In terms of FLOPs for rendering one pixel, we use 0.16 MFLOPs, whereas NeRF uses 226 MFLOPs. 

\subsection{Comparison to the State of the Art}
We compare our algorithm to state-of-the-art MPI-based methods, DeepView \cite{flynn2019deepview} and LLFF \cite{mildenhall2019local}, as well as non-MPI-based NeRF~\cite{mildenhall2020nerf} and neural scene representations (SRN) \cite{sitzmann2019srns}. We also compare to recent work, NSVF \cite{liu2020neural} in our supplementary material; however, their method focuses on object captures and is not designed to handle scenes with background due to the use of a bounded voxel grid. 
For evaluations, we use Spaces dataset from DeepView and Real Forward-Facing dataset from NeRF. Moreover, we introduce a significantly more challenging dataset, \emph{Shiny}, to test the limit of view-dependent modeling. 

\subsubsection{Results on Real Forward-Facing Dataset}
This dataset contains 8 scenes captured in real-world environments using a smartphone. The number of input images for each scene ranges from 20 to 62 images, each with a resolution of $1008\times 756$ pixels. We use the same train/test split as NeRF and evaluate our test results using 3 metrics: PSNR (Peak Signal-to-Noise Ratio, higher is better), SSIM (Structural Similarity Index Measure, higher is better) and LPIPS\cite{zhang2018unreasonable} (Learned Perceptual Image Patch Similarity, lower is better).

\begin{figure*}
    \label{fig-nerfvisually}
    \centering
    \renewcommand{\arraystretch}{3}
    \setlength\tabcolsep{1.5pt}
    \renewcommand{\arraystretch}{1}
    \ 
    \begin{tabular}{cccccc}
        
        \multirow{2}{*}[4em]{
                \shortstack{\includegraphics[width=4.5cm]{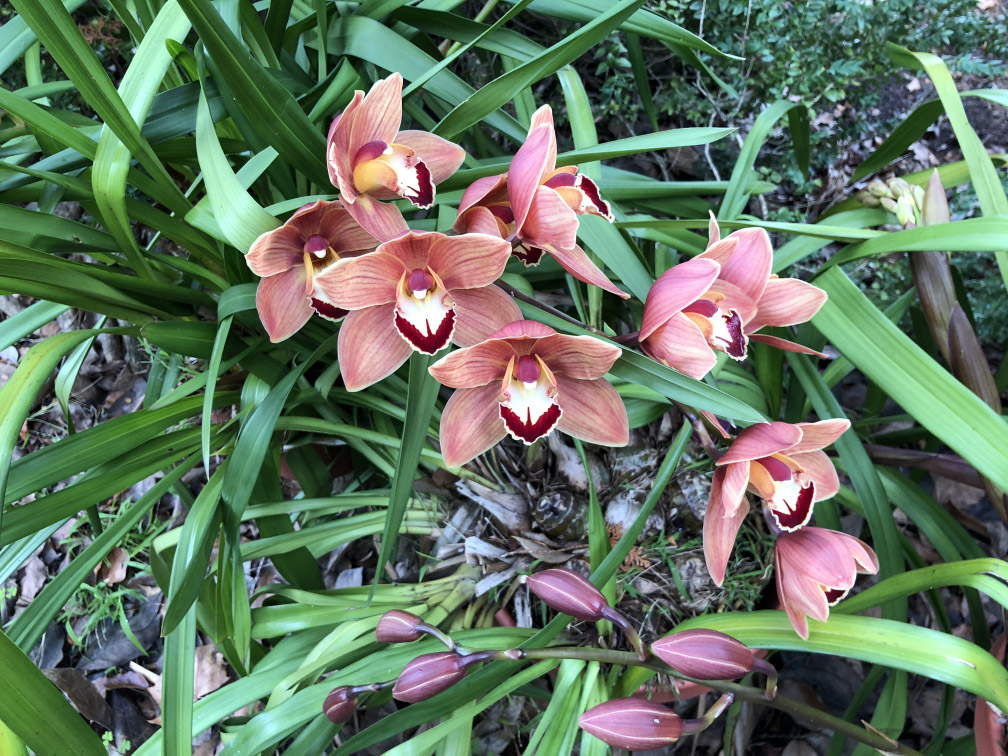} 
                \\ Orchids}} 
        & \begin{subfigure}[b]{0.10\textwidth}\includegraphics[width=\textwidth]{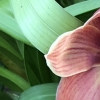}\end{subfigure}
        & \begin{subfigure}[b]{0.10\textwidth}\includegraphics[width=\textwidth]{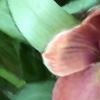}\end{subfigure}
        & \begin{subfigure}[b]{0.10\textwidth}\includegraphics[width=\textwidth]{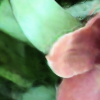}\end{subfigure} 
        & \begin{subfigure}[b]{0.10\textwidth}\includegraphics[width=\textwidth]{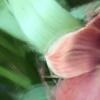}\end{subfigure}
        & \begin{subfigure}[b]{0.10\textwidth}\includegraphics[width=\textwidth]{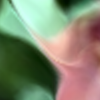}\end{subfigure}
        \\
        & \begin{subfigure}[b]{0.10\textwidth}\includegraphics[width=\textwidth]{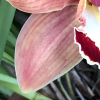}\end{subfigure}
        & \begin{subfigure}[b]{0.10\textwidth}\includegraphics[width=\textwidth]{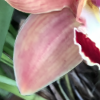}\end{subfigure}
        & \begin{subfigure}[b]{0.10\textwidth}\includegraphics[width=\textwidth]{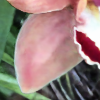}\end{subfigure} 
        & \begin{subfigure}[b]{0.10\textwidth}\includegraphics[width=\textwidth]{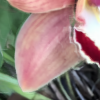}\end{subfigure}
        & \begin{subfigure}[b]{0.10\textwidth}\includegraphics[width=\textwidth]{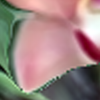}\end{subfigure}
        \\
        \addlinespace
        \multirow{2}{*}[4em]{
                \shortstack{\includegraphics[width=4.5cm]{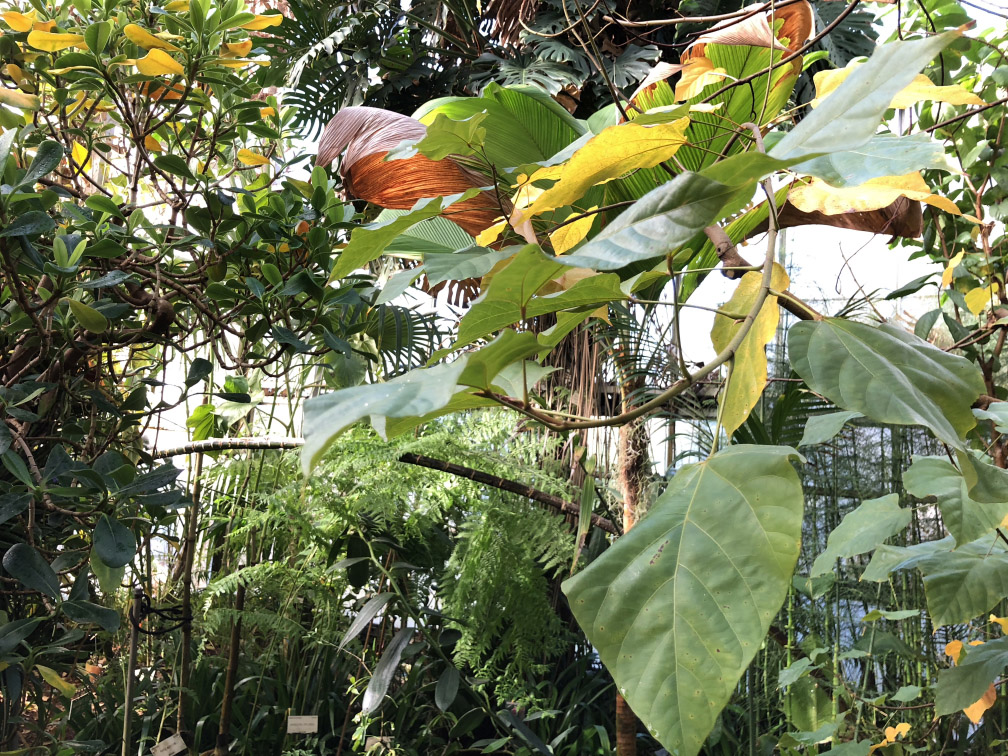} 
                \\ Leaves}} 
        & \begin{subfigure}[b]{0.10\textwidth}\includegraphics[width=\textwidth]{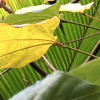}\end{subfigure}
        & \begin{subfigure}[b]{0.10\textwidth}\includegraphics[width=\textwidth]{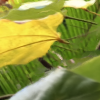}\end{subfigure}
        & \begin{subfigure}[b]{0.10\textwidth}\includegraphics[width=\textwidth]{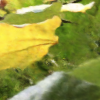}\end{subfigure} 
        & \begin{subfigure}[b]{0.10\textwidth}\includegraphics[width=\textwidth]{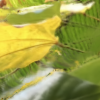}\end{subfigure}
        & \begin{subfigure}[b]{0.10\textwidth}\includegraphics[width=\textwidth]{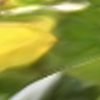}\end{subfigure}
        \\
        & \shortstack{\begin{subfigure}[b]{0.10\textwidth}\includegraphics[width=\textwidth]{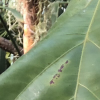}\end{subfigure}
            \\ Ground truth}
        & \shortstack{\begin{subfigure}[b]{0.10\textwidth}\includegraphics[width=\textwidth]{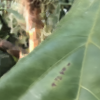}\end{subfigure}
            \\ \textbf{Ours} }
        & \shortstack{\begin{subfigure}[b]{0.10\textwidth}\includegraphics[width=\textwidth]{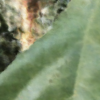}\end{subfigure} 
            \\ NeRF\cite{mildenhall2020nerf}}
        & \shortstack{\begin{subfigure}[b]{0.10\textwidth}\includegraphics[width=\textwidth]{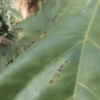}\end{subfigure}
            \\ LLFF\cite{mildenhall2019local} }
        & \shortstack{\begin{subfigure}[b]{0.10\textwidth}\includegraphics[width=\textwidth]{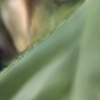}\end{subfigure}
            \\ SRN\cite{sitzmann2019srns} }
       \\
    \end{tabular}
    \caption{
    Qualitative results on test views from NeRF's real forward-facing dataset. Our method captures more complete geometry than LLFF and SRN in Orchids scene and recovers the most detail in Leaves scene.
    }
\end{figure*}

As shown in Table \ref{tab-nerf}, our method produces the highest average scores across all 3 metrics. We show scores for individual scenes in our supplementary. Note that we need to undistort the results from NeRF in order to match our calibrated testing views. By doing so, their average scores increase, and we provide both the new and original scores for reference in the supplementary material. NeRF has a higher PSNR than ours on one scene, ``Orchids,'' and upon inspection we found that our result looks distorted near the image boundary. Compared to NeRF, our results have much sharper detail and less noise in regions with uniform colors as seen in Figure \hyperref[fig-nerfvisually]{3}. The detail from LLFF is on a par with ours; however, LLFF produces jumping and warping artifacts when results are rendered as a video. SRN produces blurry results that do not look realistic for this dataset. Note that our algorithm renders state-of-the-art results more than $1000\times$ faster than NeRF, and is the first to achieve real-time over 200 FPS rendering at this quality.  
\begin{table}%
\caption{Average scores across 8 scenes in Real Forward-Facing dataset.
}
\label{tab-nerf}
\begin{minipage}{\columnwidth}
\begin{center} 
\begin{tabular}{llll}
  \toprule
  Method                          & PSNR $\uparrow$ & SSIM $\uparrow$  & LPIPS  $\downarrow$ \\ \midrule
  SRN  \cite{sitzmann2019srns}    & 21.82           & 0.744            & 0.464 \\
  LLFF \cite{mildenhall2019local} & 24.41           & 0.863            & 0.211\\
  NeRF \cite{mildenhall2020nerf}  & 26.76           & 0.883            & 0.246\\
  \textbf{NeX (Ours)}          & \textbf{27.26}  & \textbf{0.904}   & \textbf{0.178}\\
  \bottomrule
\end{tabular}
\end{center}
\end{minipage}
\end{table}

\subsubsection{Results on Shiny Dataset}
Our Shiny dataset also contains 8 scenes captured with a smartphone in a similar manner as Real Forward-Facing dataset. However, the scenes contain much more challenging view-dependent effects, such as the rainbow reflections on a CD, refraction through a liquid bottle or a magnifying glass, metallic and ceramic reflections, and sharp specular highlights on silverware, as well as detailed thin structures.

Table \ref{tab-shiny} shows that our method also outperforms NeRF on all 3 metrics on this dataset. In scene CD, our method can reproduce the rainbow reflections and the reflected image of a plastic cup on the CD, while NeRF fails to capture the reflected image, as seen in Figure \ref{fig_shiny}.  In scene Tools, our method produces a sharper image of the solder coil stand through the magnifying glass. In scene Food, our method captures the specular microgeometry of the textured ceramic plate with high fidelity. Our failure cases include the lack of sharp sparkles in the crystal candle holder in scene Food and the reflection of the tube rack in scene Lab shown in Figure \ref{fig_failure}. Currently, no other methods are able to handle extremely sharp highlights that only appear in one distinct location in each input view. 

\begin{table}%
\vspace{0.3cm}
\caption{Average scores across 8 scenes in Shiny dataset.}
\label{tab-shiny}
\begin{minipage}{\columnwidth}
\begin{center} 
\begin{tabular}{llll}
  \toprule
  Method                          & PSNR $\uparrow$ & SSIM $\uparrow$  & LPIPS  $\downarrow$ \\ \midrule
  NeRF \cite{mildenhall2020nerf}  & 25.60           & 0.851           & 0.259\\
  \textbf{NeX (Ours)}          & \textbf{26.45}  & \textbf{0.890}   & \textbf{0.165}\\
  \bottomrule
\end{tabular}
\end{center}
\end{minipage}
\vspace{-0.3cm}
\end{table}%

\subsubsection{Results on DeepView's Spaces Dataset}
Spaces dataset contains indoor and outdoor captures using 16 forward-facing cameras on a fixed rig. Each image has a resolution of $800 \times 480$. We evaluate on the same 10 scenes in Spaces dataset as in DeepView. We train our model on 12 input views, then evaluate on 4 held-out views. 
Table \ref{tab-deepview} shows a comparison between Soft3D \cite{penner2017soft}, DeepView \cite{flynn2019deepview}, and our work. Note that DeepView only estimates an MPI with 80 planes, and these scores are computed from the test images released by those papers. 
Our method produces higher average scores than DeepView on all metrics for the 12-view setup. Figure \ref{fig_shiny} shows close-up results on one of the scenes from Spaces dataset. Note that DeepView focuses on sparser input setups than ours and can produce reasonable results with 4 input views by learning from a large dataset of scenes. However, it uses the original MPI representation which handles limited view-dependent effects.

\begin{table}
\vspace{0.5em}
\caption{Average scores on Spaces dataset (12 input views).}
\label{tab-deepview}
\renewcommand{\arraystretch}{1}
\begin{center}
\vspace{-1em}
\begin{tabular}{llll} 
  \toprule
  Method                                    &PSNR$\uparrow$  & SSIM $\uparrow$  & LPIPS $\downarrow$  \\ \midrule
  Soft3D  \cite{penner2017soft}             & 31.57          & 0.964            & 0.126\\
  Deepview\cite{flynn2019deepview}          & 31.60          & 0.978            & 0.085\\
  \textbf{NeX (Ours)}  \hspace{0.5cm}       & \textbf{35.84} & \textbf{0.985}   & \textbf{0.083}\\
  \bottomrule
\end{tabular}
\end{center}
\vspace{-1.5em}
\end{table}
\begin{figure*}
    \centering
    \setlength\tabcolsep{1.5pt}
    \begin{tabular}{ccc ccc ccc}
    \multicolumn{3}{c}{ \includegraphics[trim=0 0 0 30,clip,width=5.5cm]
            {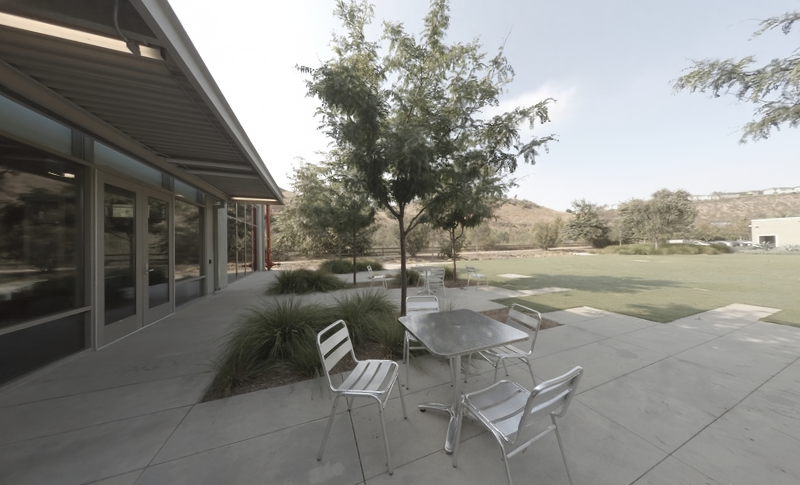} }&
    \multicolumn{3}{c}{ \includegraphics[trim=10 0 10 0,clip,width=5.5cm]
            {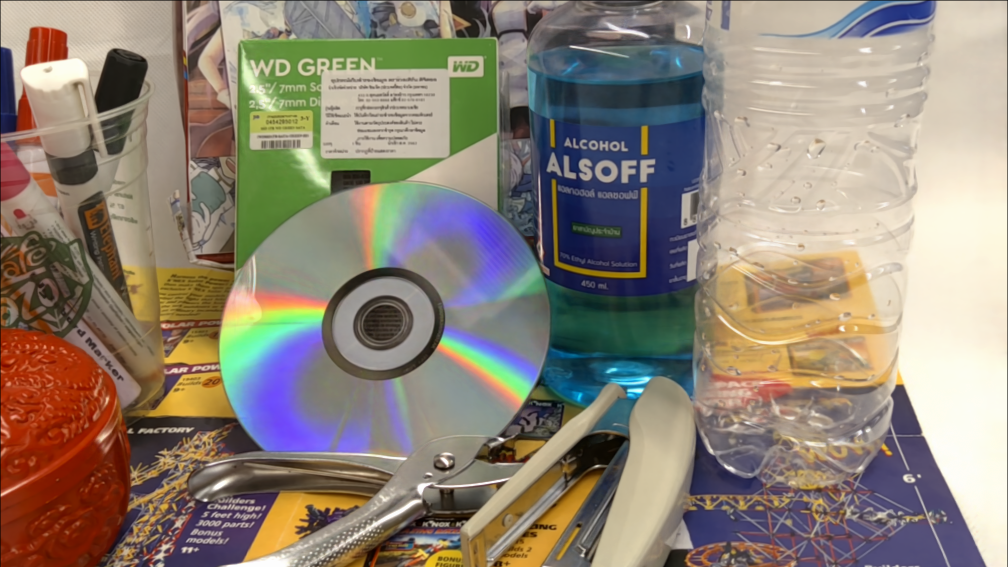} }&
    \multicolumn{3}{c}{ \includegraphics[trim=0 30 0 150,clip,width=5.5cm]
            {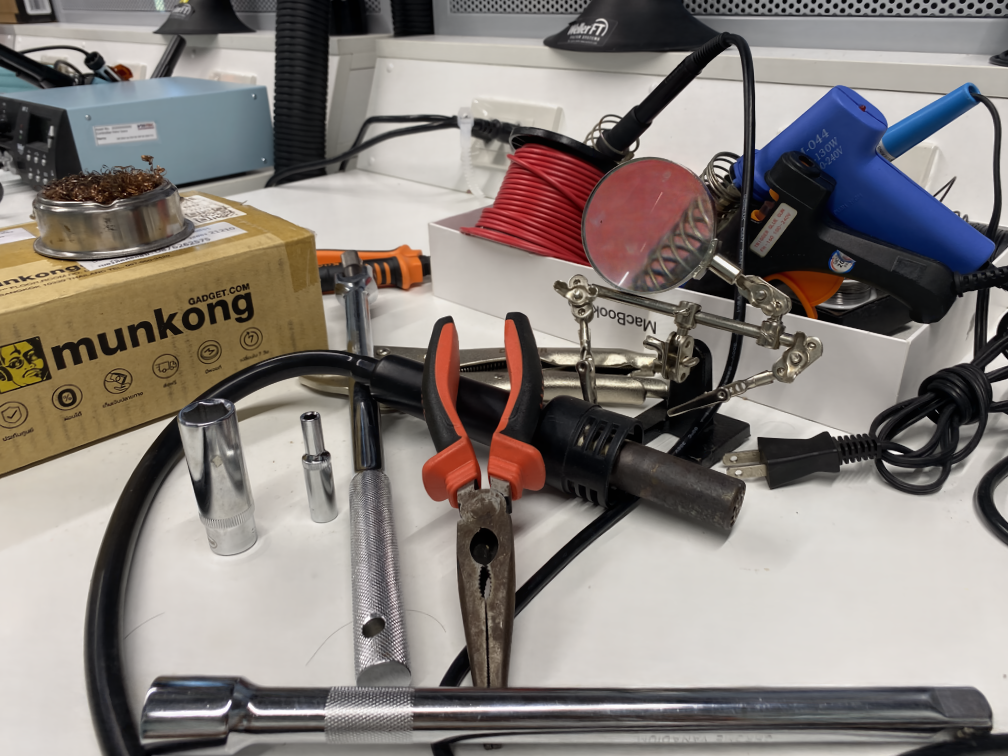} } \\

        \shortstack{ 
          \includegraphics[width=0.10\textwidth]{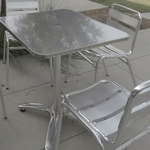}
         \\Ground truth}
        & \shortstack{ 
          \includegraphics[width=0.10\textwidth]{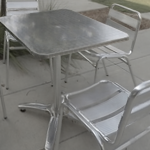}
         \\\textbf{Ours}}
        & \shortstack{
          \includegraphics[width=0.10\textwidth]{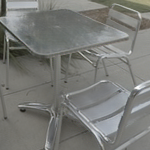}
         \\ DeepView\cite{flynn2019deepview} }
    
    &
        \shortstack{ 
          \includegraphics[width=0.10\textwidth]{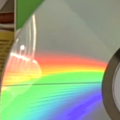}
         \\  Ground truth}
        & \shortstack{ 
          \includegraphics[width=0.10\textwidth]{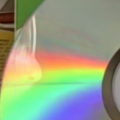}
         \\ \textbf{Ours}}
        & \shortstack{
          \includegraphics[width=0.10\textwidth]{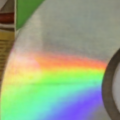}
         \\ NeRF\cite{mildenhall2020nerf} }
    &
    
        \shortstack{ 
          \includegraphics[width=0.10\textwidth]{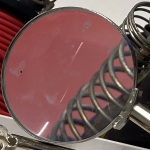}
         \\ Ground truth}
        & \shortstack{ 
          \includegraphics[width=0.10\textwidth]{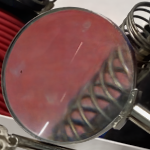}
         \\ \textbf{Ours}}
        & \shortstack{
          \includegraphics[width=0.10\textwidth]{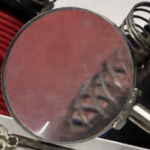}
         \\ NeRF\cite{mildenhall2020nerf} }
         
    \\[.2cm]
    
    \multicolumn{3}{c}{  (a) Spaces dataset: Scene 056 }&
    \multicolumn{3}{c}{  (b) Shiny dataset: CD }&
    \multicolumn{3}{c}{  (c) Shiny dataset: Tools } \\

    \end{tabular}
    \caption{The top row shows our rendered results. (a) Our method captures more accurate reflections on the table top. (b) Our method captures the reflected image of a plastic cup as well as the rainbow reflections, while NeRF produces a blurry and noisy result. (c) Our method produces a sharper image of the coil through the magnifying glass.}
    \label{fig_shiny}
\end{figure*}
\subsection{Ablation Studies}
We evaluate the effectiveness of our main contributions which are learned basis functions for view-dependent pixel representation and the implicit-explicit modeling strategy. For ablation studies, we train a 72-layer MPI with $M=6$ sharing scheme and test on
two scenes: ``Tools,'' which contains multiple types of view-dependent effects, and ``Crest,'' which contains high-detail patterns and thin structures from Shiny dataset. All images are in $1008\times 756$ resolution.

\subsubsection{View-dependent Modeling \& Basis Functions}\label{typeofbasis}
\begin{figure}
  \centering
  \includegraphics[scale=0.5]{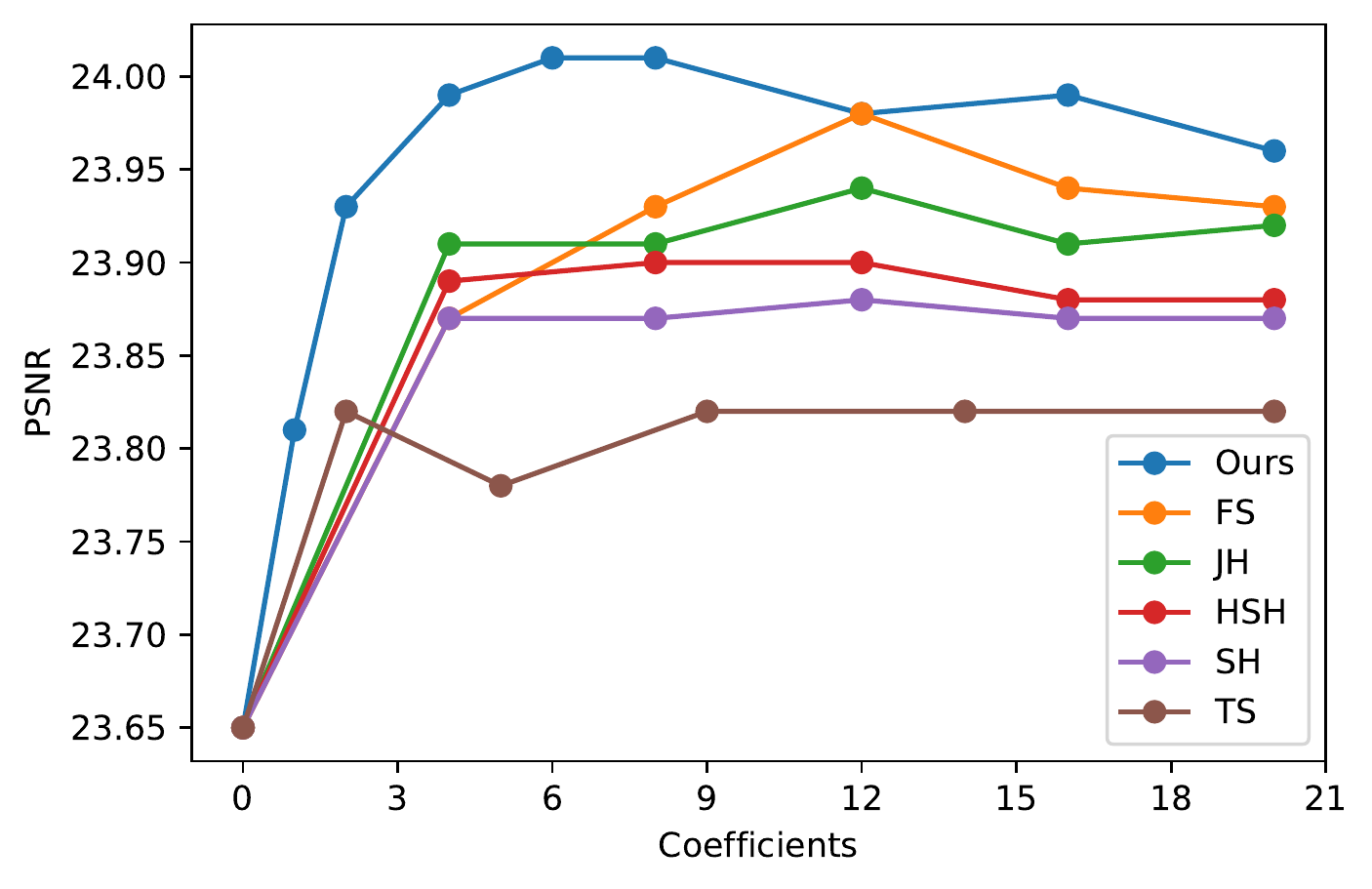}
  \caption{PSNR scores vs. the number of basis coefficients for ours (learned basis functions), FS (Fourier's series), JH (Jacobi spherical harmonics), HSH (hemispherical harmonics), SH (spherical harmonics), and TS (Taylor's series).}
  \label{PSNRvs}
\end{figure}

\begin{figure}
    \centering
    \setlength\tabcolsep{1.5pt}
    \begin{tabular}{ccc}
        \shortstack{ 
          \includegraphics[width=0.11\textwidth]{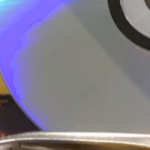}
         \\ Ground truth}
        & \shortstack{ 
          \includegraphics[width=0.11\textwidth]{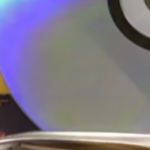}
         \\ \textbf{Ours}}
        & \shortstack{
          \includegraphics[width=0.11\textwidth]{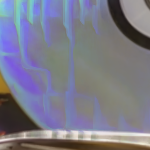}
         \\  Standard MPI}
         
    \end{tabular}
    \caption{Our MPI with view-dependent modeling can replicate the rainbow reflections on the CD, while the standard MPI breaks down completely.
    }
    \label{No_view}
\end{figure}

\hspace{8pt} 
\textbf{Number of basis coefficients:} We vary the number of basis coefficients from zero, which represents no view-dependent modeling, to 20 and show quantitative results in Figure \ref{PSNRvs}. The scores of our learned basis functions peak around 6-9 coefficients and show signs of overfiting afterward. Adding view-dependent modeling to MPI helps increase PSNR scores 
on all test scenes and significantly improves the visual quality for scenes with challenging lighting effects shown in Figure \ref{No_view}. 
\label{sec_BasisType}

\textbf{Types of basis functions:} We compare our learned basis functions to other types of basis for modeling view-dependent effects by only changing ${H_n}$ in Equation \ref{eq:basis}. We test the following basis options: Taylor Series (TS), Spherical Harmonics (SH), Hemispherical Harmonics (HSH) \cite{Gautron2004Hemi}, Jacobi Spherical Harmonics (JH), and Fourier Series (FS). 
Spherical harmonics are commonly used for representing complex illumination and are derived from Legendre polynomials. However, since our captures are mostly forward-facing, the viewing directions from which a point on a surface can be observed will only span a hemisphere. Thus, we also evaluate alternative basis functions that are more suitable for this viewing span, namely Hemi-spherical harmonics \cite{Gautron2004Hemi}, which are derived from shifted Legendre polynomials. Generalizing this further, one can derive modified spherical harmonics that target an even tighter viewing span than a hemisphere through shifted Jacobi polynomials (JH). The exact forms are shown in our supplementary material.

Figure \ref{PSNRvs} shows that our learned basis outperforms these fixed basis functions, even the ones whose viewing domains have been narrowed down, when the same number of coefficients is used. In principle, given a sufficiently expressive network, our learned basis can approximate other kinds of basis functions, if required, or reproduce higher frequencies using the same rank order, which is the number that dictates the highest frequency in those fixed basis functions.

\vspace{-1em}

\subsubsection{Implicit Function and Explicit Structure}

In this experiment, we validate our design decision that stores base color $K_0$ explicitly while modeling other parameters with implicit functions. Additionally, in Table \ref{tab:represent} we explore all 8 alternatives for representing alpha ($A$), base color ($K_0$), and view-dependent coefficients ($K_1,...,K_N$), ranging from fully implicit (Im-Im-Im) to fully explicit (Ex-Ex-Ex). Note that the fully explicit model corresponds to optimizing the rendering equation directly without any deep priors on the parameters. 
The result shows that our Im-Ex-Im outperforms other alternatives, and storing base color $K_0$ explicitly is beneficial also to other configurations regardless of the modeling choices for the alpha and coefficients. Qualitatively, our method produces significantly better detail compared to the fully implicit model and cleaner results that generalize better than the fully explicit model.


\begin{table}
\caption{Quantitative evaluation on different modeling strategies for the alpha transparency ($A$), base color ($K_0$), and view-dependent coefficients ($K_1,...,K_N$). Modeling with an explicit structure is denoted by (Ex) and with an implicit representation is denoted by (Im).}
\label{tab:represent}
\begin{minipage}{\columnwidth}
\begin{center}
\small
\begin{tabular}{c|c|c|lll}
  \toprule
  \multicolumn{3}{c|} {Method} & \multicolumn{3}{c} {Metric} \\\midrule 
  $A$ & $K_0$ & $K_1,...,K_n$  & PSNR $\uparrow$ & SSIM $\uparrow$ & LPIPS $\downarrow$  \\ \midrule 
  Ex& Ex& Ex                    & 24.57           & 0.857          & 0.292 \\
  Ex& Ex& Im                    & 24.47           & 0.854          & 0.300 \\
  Ex& Im& Ex                    & 24.55           & 0.857          & 0.296 \\
  Ex& Im& Im                    & 24.44           & 0.854          & 0.302 \\
  Im& Ex& Ex                    & 26.30           & 0.901          & 0.204 \\
  \textbf{Im}& \textbf{Ex}& \textbf{Im} & \textbf{26.32}  & \textbf{0.904} & \textbf{0.202} \\
  Im& Im& Ex                    & 25.82           & 0.883          & 0.279 \\
  Im& Im& Im                    & 25.63           & 0.878          & 0.301 \\
  \bottomrule
\end{tabular}
\end{center}
\end{minipage}
\end{table}

  \section{Limitations \& Failure Cases}
\begin{figure}
    \centering
    \setlength\tabcolsep{1.5pt}
    \begin{tabular}{cccc}
        \shortstack{ 
          \includegraphics[width=0.11\textwidth]{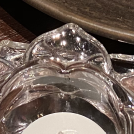}
         \\ Ground truth}
        & \shortstack{ 
          \includegraphics[width=0.11\textwidth]{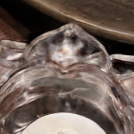}
         \\ \textbf{Ours}}
        & \shortstack{
          \includegraphics[width=0.11\textwidth]{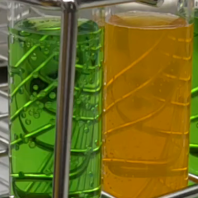}
         \\  Ground truth}
        & \shortstack{
          \includegraphics[width=0.11\textwidth]{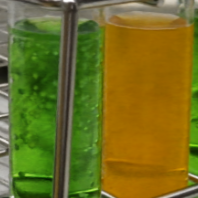}
         \\  \textbf{Ours}}
         
    \end{tabular}
    \caption{Our failure cases on a crystal candle holder in scene Food (Left) and test tubes in scene Lab (Right).
    }
    \label{fig_failure}
\end{figure}

Our model is based on MPI and thus inherits similar limitations.
When viewing our MPI from an angle too far away from the center, there will be ``stack of cards'' artifacts which can reveal individual MPI planes. Our model still cannot fully reproduce the hardest scenes in our Shiny dataset, which include effects such as light sparkles, extremely sharp highlights, or refraction through test tubes (Figure \ref{fig_failure}). Exposure differences in the training images that are not properly compensated may lead to flickering in the rendered output. Training our MPI still takes a long time and may require a higher number of input views to replicate view-dependent effects. Learning how to do this with fewer input images using a dataset of scenes or with learned optimizers \cite{flynn2019deepview} could be an interesting direction.

\section{Conclusion}
We have investigated a new approach to novel view synthesis using multiplane image (MPI) with neural basis expansion. Our representation is effective in capturing and reproducing complex view-dependent effects and efficient to compute on standard graphics hardware, thus allowing real-time rendering. Extensive studies on public datasets and our more challenging dataset demonstrate state-of-the-art quality of our approach. We believe neural basis expansion can be applied to the general problem of light field factorization and enable efficient rendering for other scene representations not limited to MPI. Our insight that some reflectance parameters and high-frequency texture can be optimized explicitly can also help recovering fine detail, a challenge faced by existing implicit neural representations.


  \section*{Acknowledgement}
This research was supported by PTT public company limited, SCB public company limited, the Program Management Unit for Human Resources \& Institutional Development, Research and Innovation Thailand (NXPO 1426293), and VISTEC. The authors also would like to thank Jeong Joon Park for useful discussions. \\
\vspace{-0.8cm}
  {
    \small
    \bibliographystyle{template/CVPR/ieee_fullname}
    \bibliography{content/bibliography.bib}
  }
  \appendix
  \newpage
  \setcounter{table}{0}
\renewcommand{\thetable}{\thesection.\arabic{table}}
\renewcommand{\thefigure}{\thesection.\arabic{figure}}
\renewcommand{\thesection}{\Alph{section}}

\section{Additional Implementation Settings}

\subsection{Image Preparation Details}
We calibrate a set of input images using an a Structure-from-Motion (SfM) algorithm in an open-source software package COLMAP  \cite{schoenberger2016sfm}. For COLMAP, we use a ``simple radial'' camera model with a single radial distortion coefficient and a shared intrinsic for all images. We use ``sift feature guided matching'' option in the exhaustive matcher step of SfM and also refine principle points of the intrinsic during the bundle adjustment.  Note that accurate camera poses and intrinsic parameters are crucial for our method, and errors in these parameters can lead to poor results. 

\subsection{Ray Sampling for Training}
During training, generating a reasonable sized output image via the rendering equation for all pixels at once is not feasible due to the memory limit on our GPU. To solve this, we only sample a subset of pixels from the entire image in each iteration of the optimization. And to facilitate the computation of image gradient needed in our loss function, if a pixel $(x, y)$ is sampled in the process, we also sample $(x+1, y)$ and $(x, y+1)$ so that the image gradients in both x and y directions can be computed through finite difference. In our implementation, we sample 2667 sets of these triplet pixels, resulting in 8001 samples. 


For evaluation, we use 3 metrics: PSNR, SSIM, and LPIPS.
Functions for computing PSNR and SSIM come from scikit-image software package, and for LPIPS, we use a VGG variant from \cite{zhang2018unreasonable} \footnote{https://github.com/richzhang/PerceptualSimilarity}.



\section{Additional Experimental Details}
\subsection{Comparison on Real Forward-Facing Dataset}
Real Forward-Facing dataset is provided by NeRF \cite{mildenhall2020nerf} and contains 8 scenes. We show 
a per-scene breakdown of the results from Table 1 in the main paper in Table \ref{tab12}. These scores from NeRF are computed from undistorted versions of their results using our estimated radial distortion parameter. We provided their original reported scores for reference in Table \ref{tab13}.
A qualitative comparison can be seen in Figure 3 in the main paper as well as in our supplementary video, which shows that our method achieves sharper fine detail.

We measured the training time on a single NVIDIA V100 with a 20-core Intel Xeon Gold 6248.
For scene Fern with 17 input photos, the training took around 18 hours.
For scene Flower with 30 input photos, the training took around 27 hours.


\begin{table*}
  \caption{Per-scene breakdown results from NeRF's Real Forward-Facing dataset.}
\begin{center}
\begin{tabular}{l|llll|llll|llll}
  \toprule
                    & \multicolumn{4}{c}{PSNR$\uparrow$}           & \multicolumn{4}{c}{SSIM$\uparrow$}                 & \multicolumn{4}{c}{LPIPS$\downarrow$} \\
                    & SRN  & LLFF & NeRF  & Our          & SRN  & LLFF & NeRF  & Our                & SRN  & LLFF & NeRF  & Our  \\ \midrule
  Fern          & 20.29 & 23.09 & 25.49 & \textbf{25.63} & 0.700 & 0.828 & 0.866 & \textbf{0.887} & 0.529 & 0.243 & 0.278 & \textbf{0.205} \\
  Flower        & 23.94 & 25.81 & 27.64 & \textbf{28.90} & 0.819 & 0.907 & 0.906 & \textbf{0.933} & 0.390 & 0.168 & 0.212 & \textbf{0.150} \\
  Fortress      & 25.70 & 29.56 & 31.34 & \textbf{31.67} & 0.816 & 0.934 & 0.941 & \textbf{0.952} & 0.494 & 0.171 & 0.166 & \textbf{0.131} \\
  Horns         & 23.15 & 25.13 & 28.02 & \textbf{28.46} & 0.801 & 0.905 & 0.915 & \textbf{0.934} & 0.479 & 0.197 & 0.258 & \textbf{0.173} \\
  Leaves        & 17.21 & 19.85 & 21.34 & \textbf{21.96} & 0.556 & 0.769 & 0.782 & \textbf{0.832} & 0.526 & 0.226 & 0.308 & \textbf{0.173} \\
  Orchids       & 16.97 & 18.73 & \textbf{20.67} & 20.42 & 0.575 & 0.703 & 0.755 & \textbf{0.765} & 0.528 & 0.308 & 0.312 & \textbf{0.242} \\
  Room          & 25.63 & 28.45 & 32.25 & \textbf{32.32} & 0.908 & 0.957 & 0.972 & \textbf{0.975} & 0.351 & 0.175 & 0.196 & \textbf{0.161} \\
  T-rex         & 21.71 & 24.67 & 27.36 & \textbf{28.73} & 0.784 & 0.903 & 0.929 & \textbf{0.953} & 0.412 & 0.204 & 0.234 & \textbf{0.192} \\
  \bottomrule
\end{tabular}
\end{center}
\bigskip\centering
\label{tab12}
\end{table*}%

\begin{table*}
  \caption{(For reference only) Original reported scores from NeRF~\cite{mildenhall2020nerf} where test images are not undistorted.}
\begin{center}
\begin{tabular}{l|lll|lll|lll}
  \toprule
                    & \multicolumn{3}{c}{PSNR$\uparrow$}           & \multicolumn{3}{c}{SSIM$\uparrow$}                 & \multicolumn{3}{c}{LPIPS$\downarrow$} \\
                    & SRN  & LLFF & NeRF           & SRN  & LLFF & NeRF                 & SRN  & LLFF & NeRF   \\ \midrule
  Fern          & 21.37 & 21.37 & 25.17    & 0.822 & 0.887 & 0.932    & 0.459 & 0.247 & 0.280   \\
  Flower        & 24.63 & 25.46 & 27.40    & 0.916 & 0.935 & 0.941    & 0.288 & 0.174 & 0.219   \\
  Fortress      & 26.63 & 29.40 & 31.16    & 0.838 & 0.957 & 0.962    & 0.453 & 0.173 & 0.171   \\
  Horns         & 24.33 & 24.70 & 27.45    & 0.921 & 0.941 & 0.951    & 0.376 & 0.193 & 0.268   \\
  Leaves        & 18.24 & 19.52 & 20.92    & 0.822 & 0.877 & 0.904    & 0.440 & 0.216 & 0.316   \\
  Orchids       & 17.37 & 18.52 & 20.36    & 0.746 & 0.775 & 0.852    & 0.467 & 0.313 & 0.321   \\
  Room          & 28.42 & 28.42 & 32.70    & 0.950 & 0.978 & 0.978    & 0.240 & 0.155 & 0.178   \\
  T-rex         & 22.87 & 24.15 & 26.80    & 0.916 & 0.935 & 0.960    & 0.298 & 0.222 & 0.249   \\
  \bottomrule
\end{tabular}
\end{center}
\bigskip\centering
\label{tab13}
\end{table*}%

\subsection{Comparison on Shiny Dataset} 
Our own dataset, Shiny, consists of 8 scenes with more challenging view-dependent effects compared to Real Forward-Facing dataset. 
Table \ref{tab2} shows the image resolution and number of images for each scene. To generate results for NeRF, we use the code implemented by the authors\footnote{https://github.com/bmild/nerf} using TensorFlow and train on each scene with their default setting for 200k iterations.

We show a per-scene breakdown of the results from Table 2 in the main paper in Table \ref{tab2-1}.
Our approach achieves better performance than NeRF on all metrics in all scenes. A full visual comparison is provided in our supplementary webpage.

\begin{table}[b]
  \caption{Image resolution and the number of images for each scene in our Shiny dataset. For most scenes, only 20-50 images are enough to produce good results. However, scenes with complex view-dependent effects like CD require more images.}
\begin{center}
\vspace{-0.5cm}
\begin{tabular}{l|cc}
  \toprule
                     & image resolution & number of images         \\ \midrule
  CD          & 1008$\times$567 & 307     \\
  Tools        & 1008$\times$756 & 58      \\
  Crest       & 1008$\times$756 & 50      \\
  Seasoning      & 1008$\times$756 & 45   \\
  Food        & 1008$\times$756 & 49      \\
  Giants       & 1008$\times$756 & 32    \\
  Lab         & 1008$\times$567 & 303     \\
  Pasta       & 1008$\times$756 & 35     \\
  \bottomrule
\end{tabular}
\end{center}
\bigskip\centering
\label{tab2}
\vspace{-0.5cm}
\end{table}%

\begin{table}[b]
  \caption{Per-scene breakdown results on our Shiny dataset. 
  }
\begin{center}
\vspace{-0.5cm}
\resizebox{\columnwidth}{!}{%
\begin{tabular}{l|ll|ll|ll}
  \toprule
                    & \multicolumn{2}{c}{PSNR$\uparrow$}           & \multicolumn{2}{c}{SSIM$\uparrow$}                 & \multicolumn{2}{c}{LPIPS$\downarrow$} \\
                     & NeRF & Ours           & NeRF & Ours                  & NeRF & Ours   \\ \midrule
  CD          & 30.14 & \textbf{31.43}     & 0.937 & \textbf{0.958}     & 0.206 & \textbf{0.129}    \\
  Tools        & 27.54 & \textbf{28.16}     & 0.938 & \textbf{0.953}     & 0.204 & \textbf{0.151}    \\
  Crest       & 20.30 & \textbf{21.23}     & 0.670 & \textbf{0.757}     & 0.315 & \textbf{0.162}    \\
  \small Seasoning      & 27.79 & \textbf{28.60}     & 0.898 & \textbf{0.928}     & 0.276 & \textbf{0.168}    \\
  Food        & 23.32 & \textbf{23.68}     & 0.796 & \textbf{0.832}     & 0.308 & \textbf{0.203}    \\
  Giants       & 24.86 & \textbf{26.00}     & 0.844 & \textbf{0.898}     & 0.270 & \textbf{0.147}    \\
  Lab         & 29.60 & \textbf{30.43}     & 0.936 & \textbf{0.949}     & 0.182 & \textbf{0.146}    \\
  Pasta       & 21.23 & \textbf{22.07}     & 0.789 & \textbf{0.844}     & 0.311 & \textbf{0.211}    \\
  \bottomrule
\end{tabular}}
\end{center}
\bigskip\centering
\label{tab2-1}
\end{table}%

\begin{figure*}
    \centering
    \setlength\tabcolsep{1.5pt}
    \begin{tabular}{ccc}
        \shortstack{ 
          \begin{overpic}[width=5.5cm]{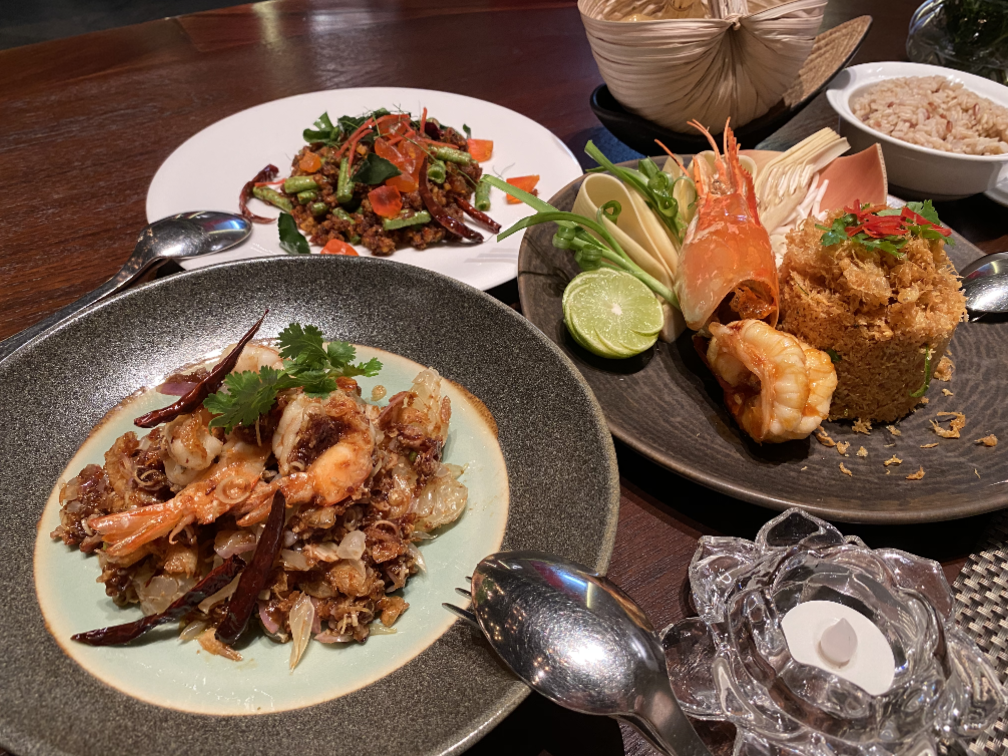}
             \put(3,3){\includegraphics[width=1.5cm,frame]{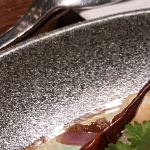}}  
          \end{overpic} }
        & \shortstack{ 
          \begin{overpic}[width=5.5cm]{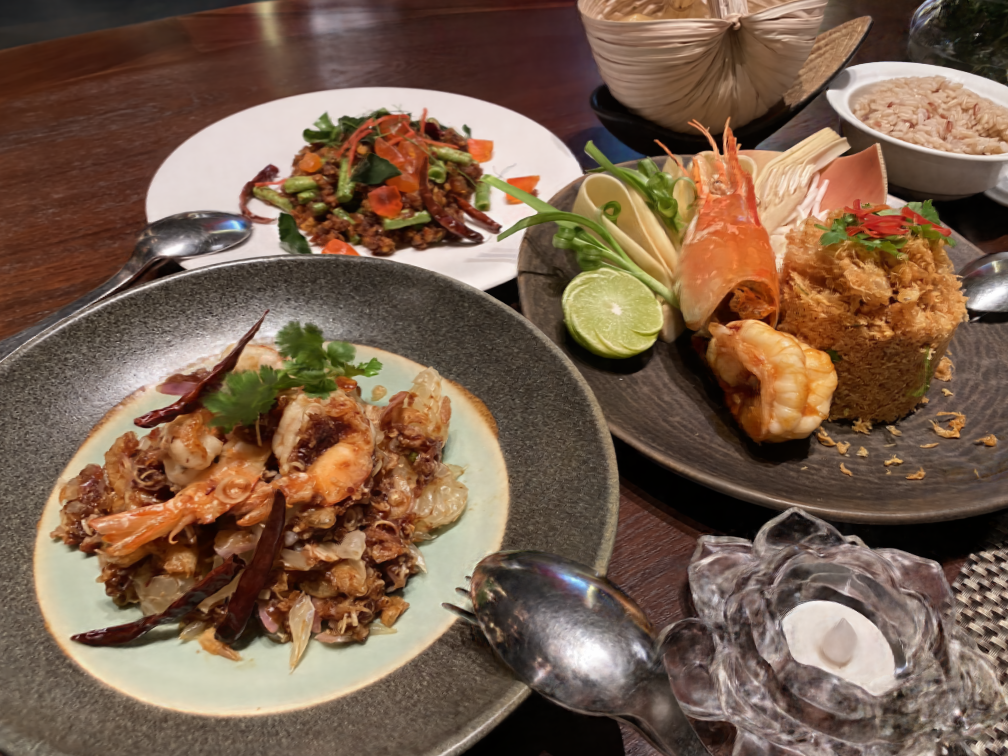}
             \put(3,3){\includegraphics[width=1.5cm,frame]{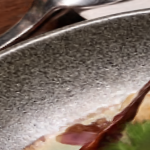}}  
          \end{overpic} }
        & \shortstack{
          \begin{overpic}[width=5.5cm]{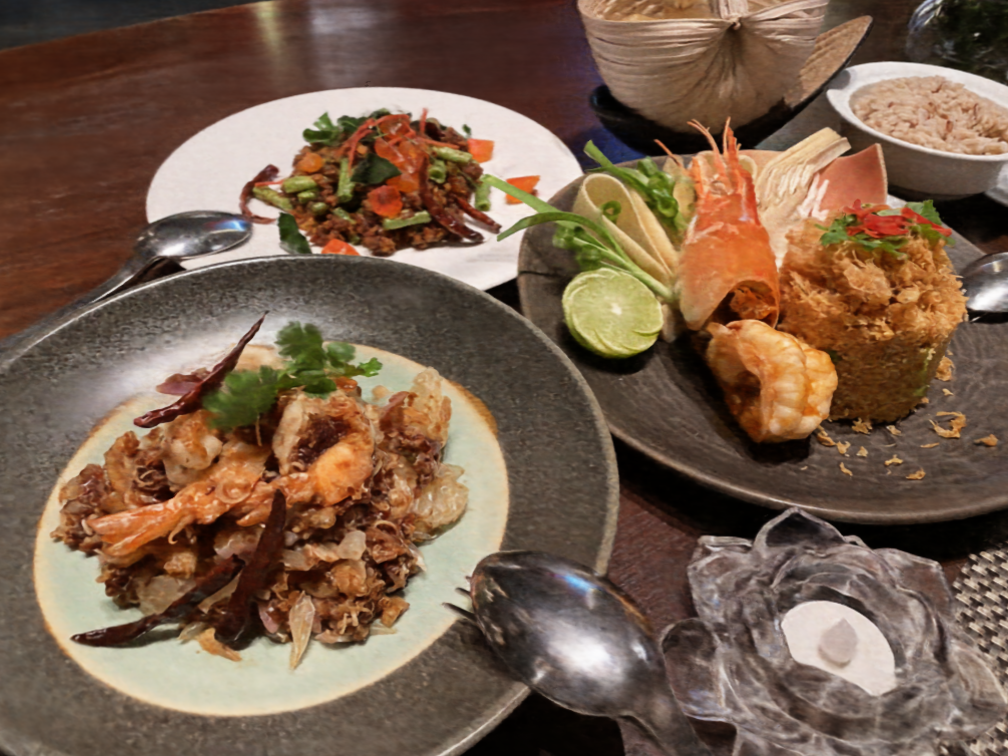}
             \put(3,3){\includegraphics[width=1.5cm,frame]{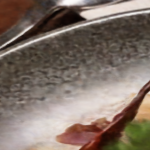}}  
          \end{overpic}  }
        \\
        \shortstack{ 
          \begin{overpic}[width=5.5cm]{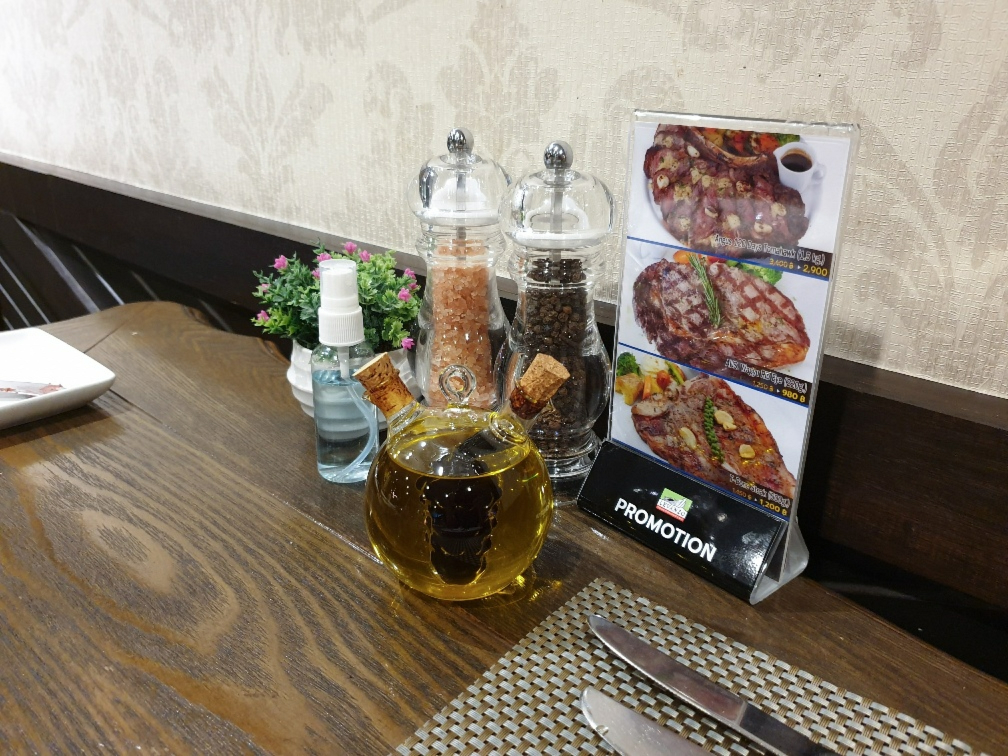}
             \put(3,3){\includegraphics[width=1.5cm,frame]{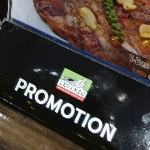}}  
          \end{overpic} 
          \\  Ground truth}
        & \shortstack{ 
          \begin{overpic}[width=5.5cm]{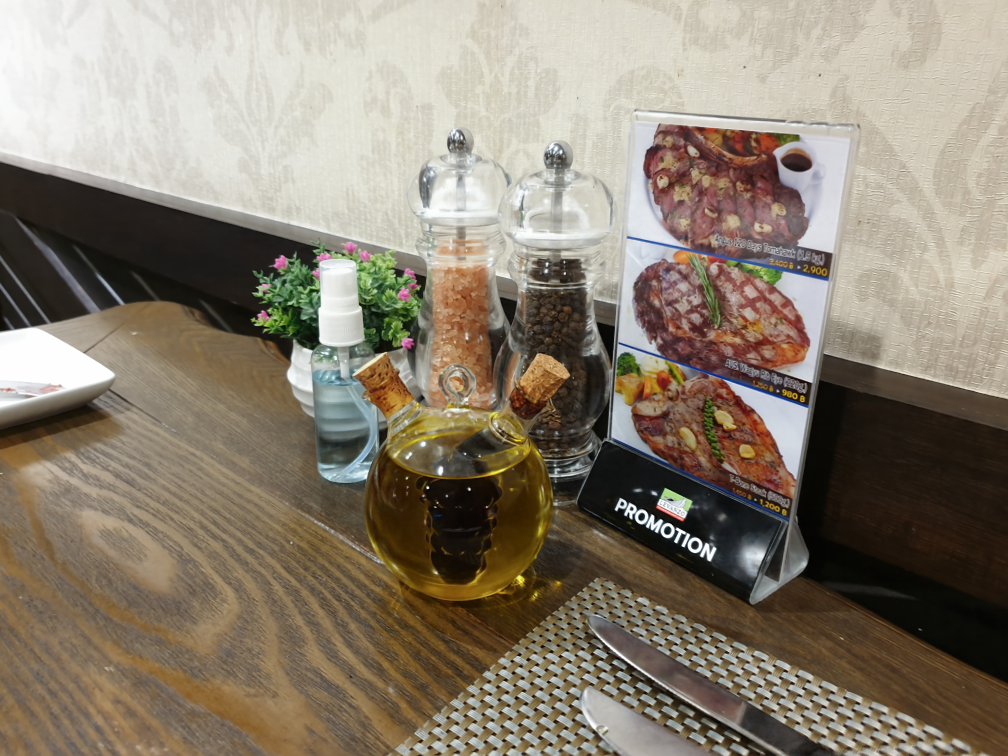}
             \put(3,3){\includegraphics[width=1.5cm,frame]{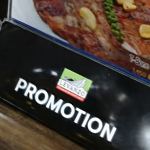}}  
          \end{overpic} 
          \\  Nex (Ours)}
        & \shortstack{
          \begin{overpic}[width=5.5cm]{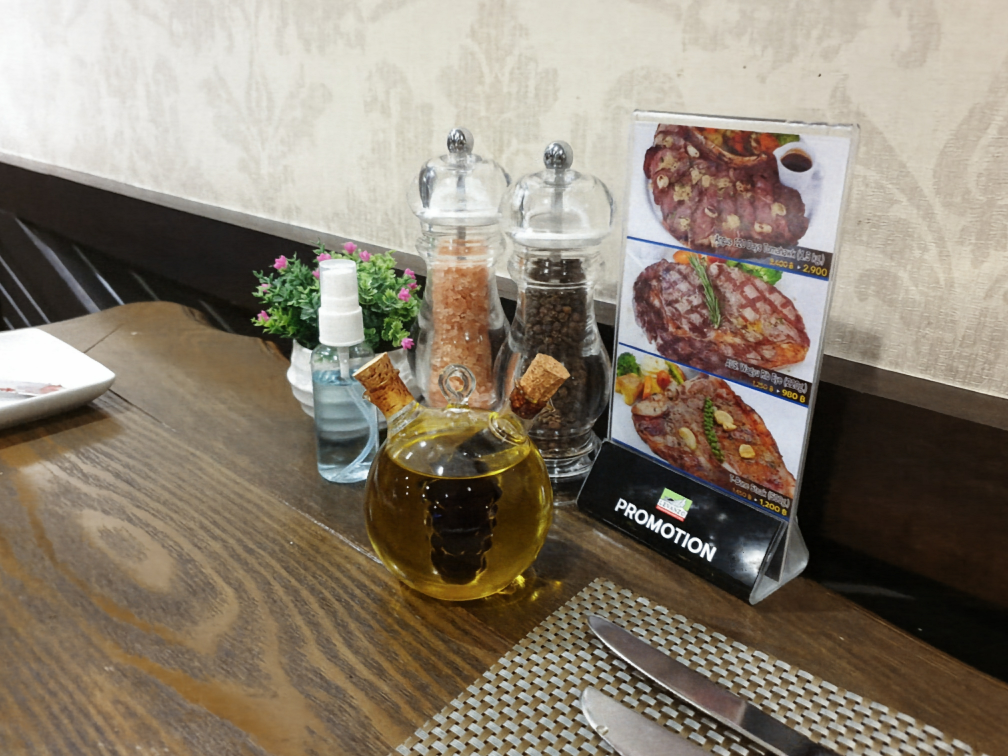}
             \put(3,3){\includegraphics[width=1.5cm,frame]{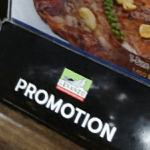}}  
          \end{overpic}  
          \\  NeRF}
    \end{tabular}
    \caption{A qualitative comparison on Shiny dataset between ground truth(left), NeX (center), and NeRF\cite{mildenhall2020nerf} (right). A full comparison on all scenes can be found in our supplementary webpage}
    \label{fig_shiny}
\end{figure*}

\subsection{Comparison on Spaces Dataset} 
The authors of DeepView have not made their code publicly available, but they have released their output results. So,  we run our algorithm on their Spaces dataset and compare our results to theirs.
Table \ref{tab3} shows a per-scene breakdown of the results from Table 3 in the main paper. A full visual comparison is provided in our supplementary webpage.

\begin{table*}[h]
  \caption{Per-scene breakdown results on Spaces dataset.}
\begin{center}
\begin{tabular}{l|ccc|ccc|ccc}
  \toprule
                & \multicolumn{3}{c}{PSNR$\uparrow$}           & \multicolumn{3}{c}{SSIM$\uparrow$}                 & \multicolumn{3}{c}{LPIPS$\downarrow$} \\
                & Soft3D  & DeepView & Ours      & Soft3D  & DeepView & Ours         & Soft3D  & DeepView & Ours   \\ \midrule
  scene000     & 32.66 & 32.54 & \textbf{37.61}     & 0.971 & 0.983 & \textbf{0.989}     & 0.093 & 0.059 & \textbf{0.049}   \\
  scene009     & 31.46 & 31.07 & \textbf{35.40}     & 0.962 & 0.972 & \textbf{0.981}     & 0.123 & 0.091 & \textbf{0.080}   \\
  scene010     & 32.94 & 31.22 & \textbf{37.61}     & 0.973 & 0.979 & \textbf{0.989}     & 0.137 & 0.095 & \textbf{0.095}   \\
  scene023     & 31.52 & 31.14 & \textbf{35.69}     & 0.969 & 0.978 & \textbf{0.986}     & 0.142 & 0.102 & \textbf{0.098}   \\
  scene024     & 33.88 & 33.15 & \textbf{37.77}     & 0.978 & 0.983 & \textbf{0.989}     & 0.119 & \textbf{0.081} & 0.090   \\
  scene052     & 30.08 & 30.22 & \textbf{34.02}     & 0.947 & 0.971 & \textbf{0.979}     & 0.119 & 0.081 & \textbf{0.076}   \\
  scene056     & 30.64 & 31.04 & \textbf{34.77}     & 0.956 & 0.975 & \textbf{0.981}     & 0.141 & 0.087 & \textbf{0.087}   \\
  scene062     & 32.56 & 32.07 & \textbf{35.34}     & 0.969 & 0.980 & \textbf{0.984}     & 0.151 & \textbf{0.098} & 0.121   \\
  scene063     & 29.72 & 32.72 & \textbf{35.44}     & 0.952 & 0.979 & \textbf{0.987}     & 0.122 & 0.078 & \textbf{0.073}   \\
  scene073     & 30.28 & 30.85 & \textbf{34.81}     & 0.960 & 0.977 & \textbf{0.986}     & 0.111 & 0.073 & \textbf{0.065}   \\
  \bottomrule
\end{tabular}
\end{center}
\bigskip\centering
\label{tab3}
\end{table*}%


\begin{figure*}
    \centering
    \setlength\tabcolsep{1.5pt}
    \begin{tabular}{ccc}
        \shortstack{
         \includegraphics[width=5.5cm]{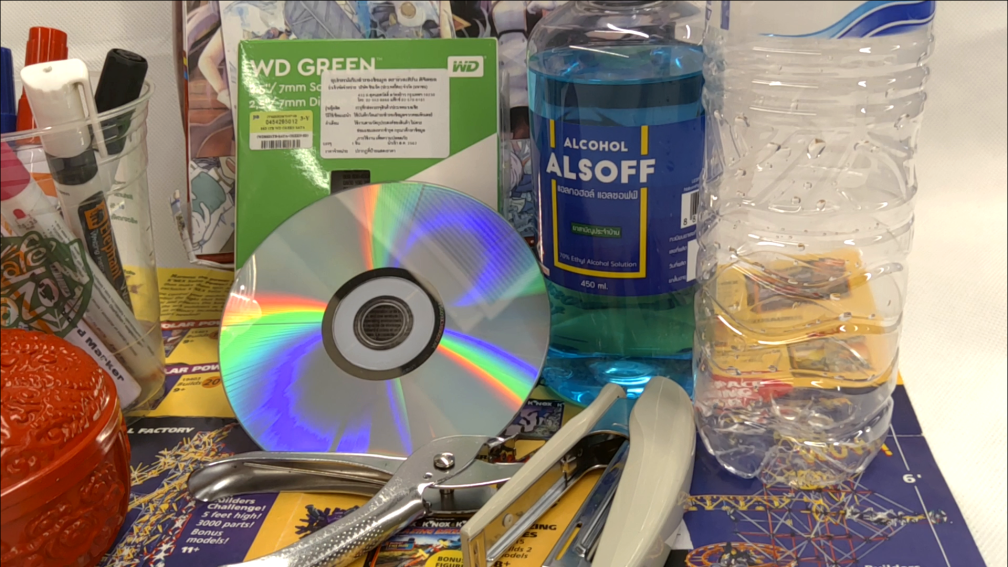}
         \\  Ground truth}
         \shortstack{
         \includegraphics[width=5.5cm]{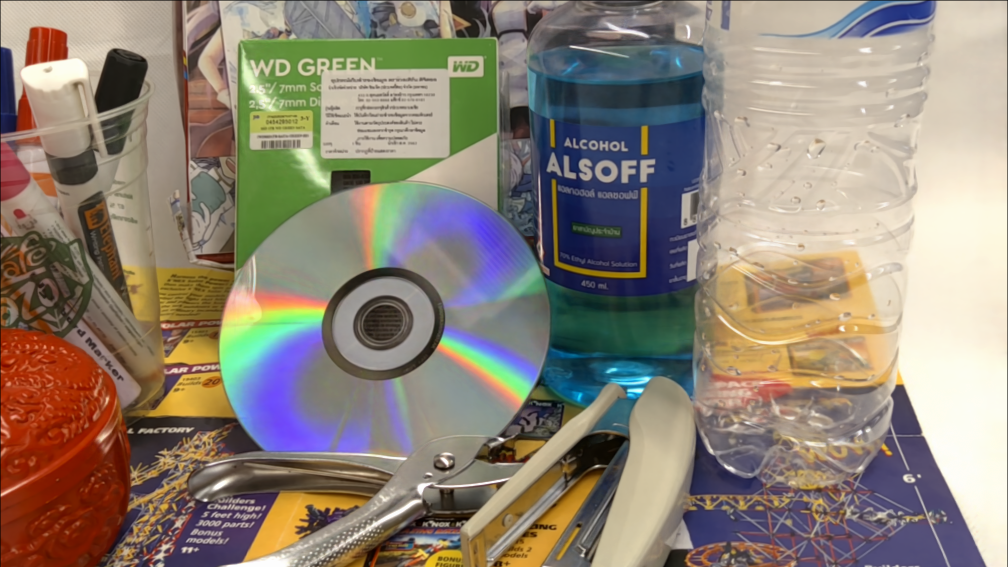}
         \\  Ours}
         \shortstack{
         \includegraphics[width=5.5cm]{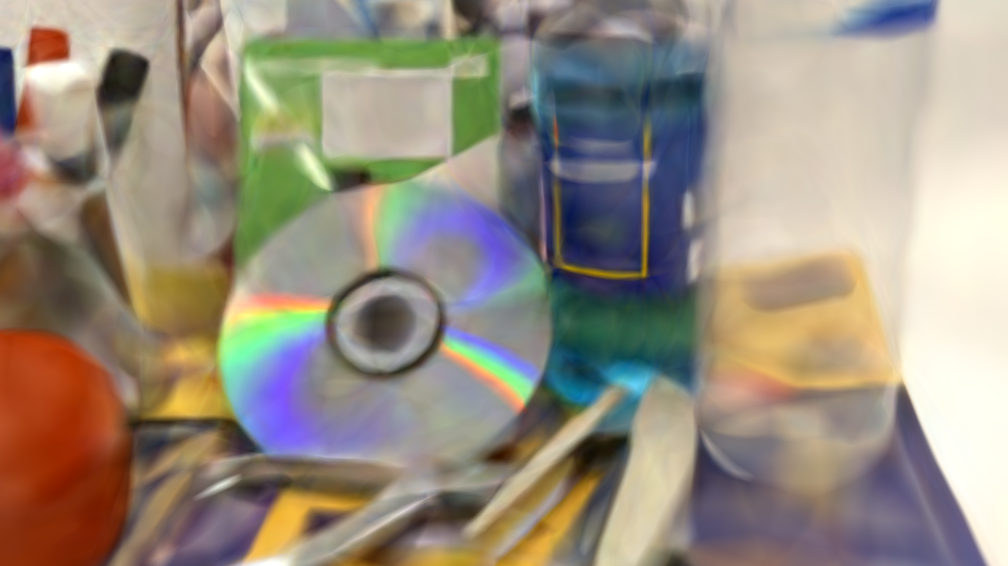}
         \\  NSVF}
    \end{tabular}
    \caption[]{A qualitative comparison on scene CD between ground truth (left), NeX (center), and NSVF~\cite{liu2020neural} (right).  
    We use NSVF code open-sourced by the authors\footnote{https://github.com/facebookresearch/NSVF}.
    NSVF does not perform well for this problem setup because it focuses on object captures where a bounding volume can be tightly defined.
    }
    \label{fig_nsvf}
\end{figure*}

\subsection{Details for Types of Basis Ablation Study}
In Section 4.3.2, we evaluate our algorithm using different sets of basis functions.
The experiment is done by changing the neural basis $\vec{H}_\phi$ in Algorithm 1 to other kinds of basis functions such as $\vec{H}_{FS}, \vec{H}_{TS}$ and $\vec{H}_{SH}$. 

Our Fourier's basis is similar to the positional encoding used in NeRF \cite{mildenhall2020nerf} and can be computed by:
\begin{align} 
   { \vec{H}_{FS}(v) = \scriptstyle \left[\cos(2^{-1}\pi v_x), \sin(2^{-1}\pi v_x), ..., \cos(2^{N}\pi v_y), \sin(2^{N}\pi v_y)\right]}.
\end{align}
For forward-facing scenarios, the viewing angle $v$ only covers a hemi-sphere. So, $v_z$ can be fully determined from $v_x$ and $v_y$ through $v_z = \sqrt{1-v_x^2-v_y^2}$, and we can parameterize the viewing angle with just $v_x$ and $v_y$ and define the FS basis only on these two parameters.

To calculate other basis functions used in Section 4.3.1, let the following complex-valued functions $K_{a,b}^{(m)}$ and $P_{a,b}^{(m)}$ be defined as:
\begin{align}
    K_{a,b}^{(m)}(v) =& \scriptstyle \left(\left( \frac{v_x}{1-a}\sqrt{\frac{v_z-a}{v_z+1}} \right) + \left(\frac{v_y}{1-a}\sqrt{\frac{v_z-a}{v_z+1}}\right) i\right)^m \\
    P_{a,b}^{(m)}(v) =& \scriptstyle \frac{ v_z -1 }{1-a} + \frac{m+1}{b+2m+2}
\end{align}
The general form of the set of basis functions is:
\begin{align}
\begin{split}
    \vec{H}(v) =& \Big[\text{Re}(K_{a,b}^{(2^0)}(v)),\\                     &\ \ \text{Im}(K_{a,b}^{(2^0)}(v)),\\
                &\ \ \text{Re}(P_{a,b}^{(2^0)}(v)\cdot K_{a,b}^{(2^0)}(v)), \\ & \ \ \text{Im}(P_{a,b}^{(2^0)}(v)\cdot K_{a,b}^{(2^0)}(v)), \\
                &\ \  ...,\\
                &\ \   \text{Re}(K_{a,b}^{(2^N)}(v)), \\ &\ \ \text{Im}(K_{a,b}^{(2^N)}(v)),  \\
                &\ \  \text{Re}(P_{a,b}^{(2^N)}(v)\cdot K_{a,b}^{(2^N)}(v)),\\ &\ \ \text{Im}(P_{a,b}^{(2^N)}(v)\cdot K_{a,b}^{(2^N)}(v))\Big]
\end{split}
\end{align}
where if $a=-1, b=0$, then it reduces to the spherical harmonics basis (SH). 
If $a=0, b=0$, then it reduces to the hemispherical harmonics basis (HSH) \cite{Gautron2004Hemi}.
For Jacobi basis (JH), we set $a=\cos(45^{\circ})=1/\sqrt{2}$ and $b=2$.

Here are examples of the first five terms for each basis that we use in Section 4.3.1:
\begin{align}
\begin{split}
    \vec{H}_{SH}(v) = &\scriptstyle [v_x/2, v_y/2, v_z v_x /2 , v_z v_y / 2, (v_x^2-v_y^2)/4, ...]       \\
    \vec{H}_{HSH}(v) =&\scriptstyle [v_x \sqrt{\frac{v_z}{v_z+1}}, v_y\sqrt{\frac{v_z}{v_z+1}},v_x\frac{2v_z-1}{2} \sqrt{\frac{v_z}{v_z+1}},    \\
                      &\scriptstyle  v_y\frac{2v_z-1}{2} \sqrt{\frac{v_z}{v_z+1}}, \frac{v_x^2-v_y^2}{(1-a)^2}\frac{v_z}{v_z-1},...]       \\
    \vec{H}_{JH}(v) = &\scriptstyle [v_x\sqrt{\frac{v_z-a}{v_z+1}}, v_y\sqrt{\frac{v_z-a}{v_z+1}}, v_x\left(\frac{v_z-1}{z-a}+\frac{2}{b+4}\right) \sqrt{\frac{v_z}{v_z+1}},\\
                      &\scriptstyle v_y\left(\frac{v_z-1}{z-a}+\frac{2}{b+4}\right) \sqrt{\frac{v_z}{v_z+1}}, \frac{v_x^2-v_y^2}{(1-a)^2}\frac{v_z}{v_z-1}, ...]
\end{split}
\end{align}

\begin{figure}
  \centering
  \includegraphics[scale=0.6]{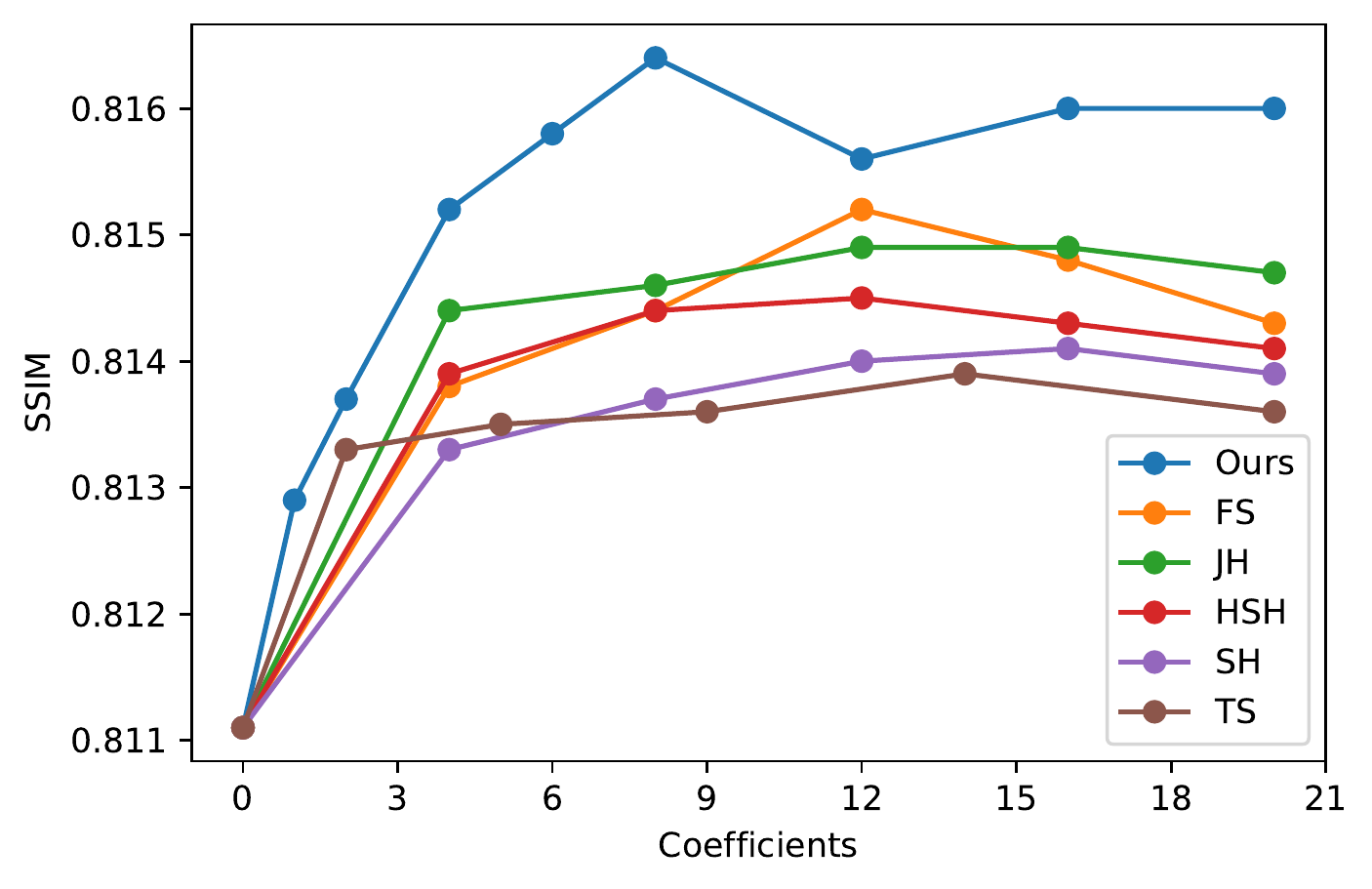}
  \caption{Number of coefficients versus SSIM score (higher is better)}
  \label{SSIMvs}
\end{figure}

\begin{figure}
  \centering
  \includegraphics[scale=0.6]{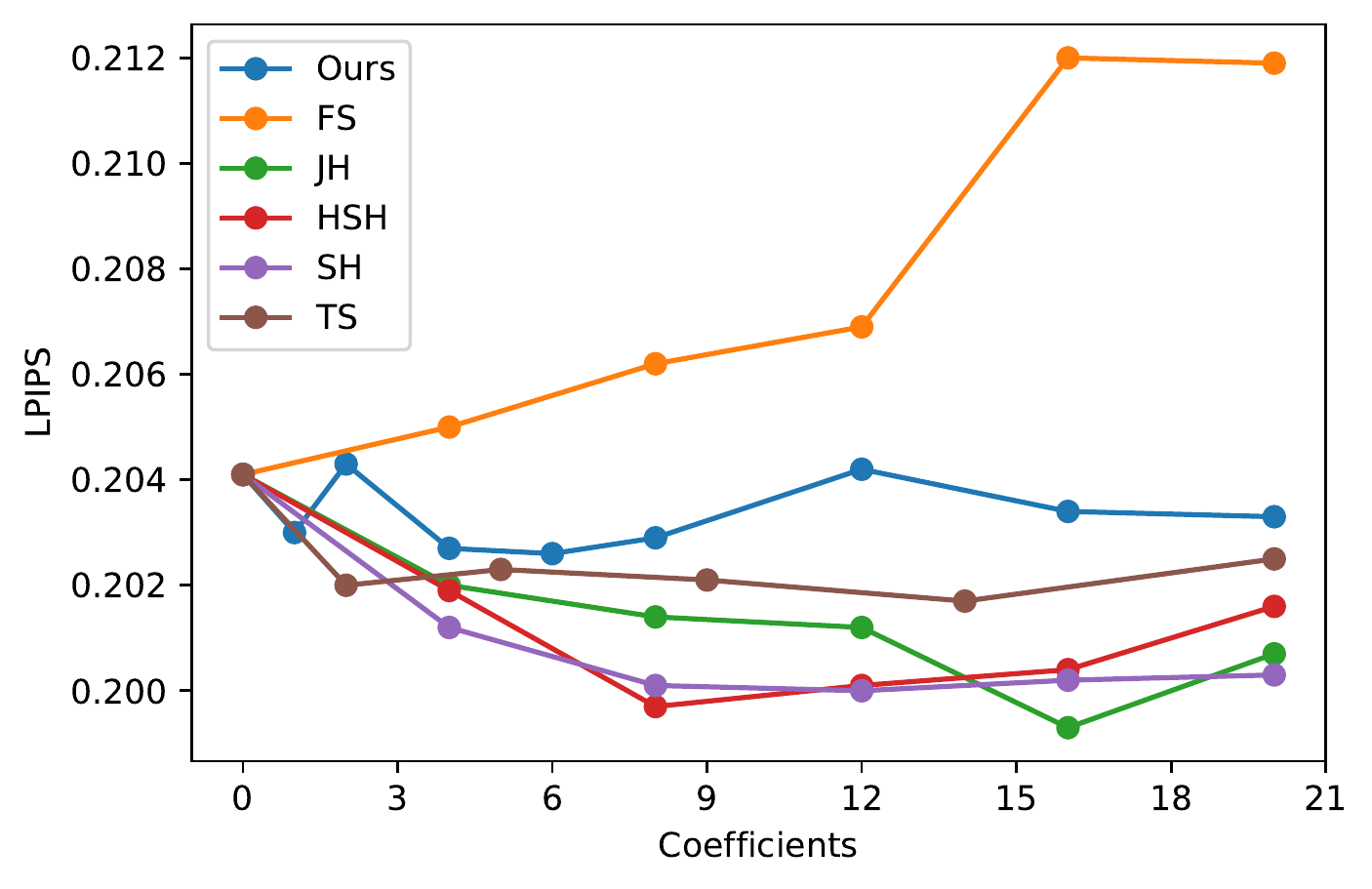}
  \caption{Number of coefficients versus LPIPS score (lower is better)}
  \label{LPIPSvs}
\end{figure}

Figure 5 in the main paper already shows PSNR scores of these basis functions. SSIM and LPIPS scores from the same experiment are shown in figure \ref{SSIMvs} and \ref{LPIPSvs} respectively.

\end{document}